\documentclass[10pt,twocolumn,letterpaper]{article}
\usepackage[pagenumbers]{cvpr}

\makeatletter
\renewcommand{\paragraph}{%
  \@startsection{paragraph}{4}%
  {\z@}{1ex \@plus .5ex \@minus .2ex}{-1em}%
  {\normalfont\normalsize\bfseries}%
}
\makeatother

\definecolor{custom_red}{RGB}{231,111,81}
\definecolor{custom_green}{RGB}{42,157,143}
\definecolor{custom_dark}{RGB}{38,70,83}
\definecolor{custom_yellow}{RGB}{233,196,106}
\definecolor{custom_orange}{RGB}{244,162,97}
\definecolor{cvprblue}{rgb}{0.21,0.49,0.74}
\definecolor{softblue}{RGB}{226,239,249}
\definecolor{softgreen}{RGB}{229,244,232}
\definecolor{softorange}{RGB}{252,237,216}
\definecolor{softgray}{RGB}{242,242,242}
\definecolor{softpurple}{RGB}{238,232,247}
\definecolor{darkblue}{RGB}{32,76,112}
\definecolor{darkgreen}{RGB}{36,94,62}
\definecolor{darkorange}{RGB}{145,78,24}
\definecolor{darkred}{RGB}{128,38,48}

\newcommand{\method}{WIS3R\xspace}
\newcommand{\iithreed}{I23D\xspace}

\newcommand{\dg}{^\circ}

\usepackage{graphicx}
\usepackage{amsmath,amssymb,mathtools}
\usepackage{booktabs,multirow,array,tabularx,makecell}
\usepackage{xcolor}
\usepackage{xspace}
\usepackage{microtype}
\usepackage{enumitem}
\usepackage{pifont}
\usepackage{cuted}
\usepackage{caption}
\usepackage{float}
\usepackage{balance}
\usepackage{url}
\usepackage{comment}
\usepackage{bm}
\usepackage{bbm}
\usepackage{colortbl}

\def\vone{{\bm{1}}}
\def\vmu{{\bm{\mu}}}
\def\vtheta{{\bm{\theta}}}

\def\vc{{\bm{c}}}
\def\vd{{\bm{d}}}

\def\vh{{\bm{h}}}

\def\vm{{\bm{m}}}

\def\vr{{\bm{r}}}
\def\vs{{\bm{s}}}
\def\vt{{\bm{t}}}
\def\vu{{\bm{u}}}
\def\vv{{\bm{v}}}

\def\vx{{\bm{x}}}
\def\vy{{\bm{y}}}
\def\vz{{\bm{z}}}
\def\vtau{{\bm{\tau}}}
\def\vdelta{{\bm{\delta}}}
\def\vell{{\bm{\ell}}}

\def\mF{{\bm{F}}}

\def\mI{{\bm{I}}}

\def\mK{{\bm{K}}}
\def\mL{{\bm{L}}}
\def\mM{{\bm{M}}}

\def\mT{{\bm{T}}}

\def\mW{{\bm{W}}}
\def\mX{{\bm{X}}}

\def\mSigma{{\bm{\Sigma}}}

\definecolor{cvprblue}{rgb}{0.21,0.49,0.74}
\usepackage[pagebackref,breaklinks,colorlinks,allcolors=cvprblue]{hyperref}

\def\confName{CVPR}
\def\confYear{2027}

\title{\method: Watermarking Images with Zero-Shot Survival across 3D Reconstructors}

\author{Akis Iochanas$^{1}$ \qquad Nikitas Chatzis$^{1}$ \qquad Marios Loizou$^{1,2}$ \qquad Evangelos Kalogerakis$^{1}$\\
$^{1}$Technical University of Crete \qquad $^{2}$CYENS Centre of Excellence}

\begin{document}
\maketitle

\begin{strip}
\centering
\vspace{-14mm}
\includegraphics[width=\textwidth]{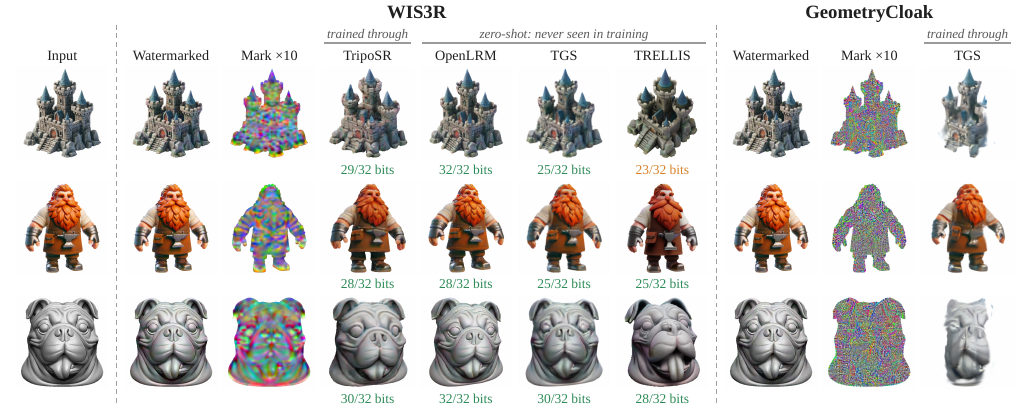}
\captionsetup{hypcap=false}
\vspace{-6mm}
\captionof{figure}{\textbf{\method} embeds a hidden 32-bit key in the object image. 
Each image-to-3D model reconstructs a 3D asset from the published image, and the key is read back blind from \textbf{one render} of that asset. 
Under each render is the number of its 32 bits matching the key. 
\method was trained on TripoSR alone, the rest of the models are never seen in training (\textbf{zero-shot}). 
Geometry Cloak~\cite{song2024geometrycloak} instead deliberately alters TGS's reconstructed geometry to expose a visible watermark, deterring faithful reconstruction rather than preserving the asset for reuse.}
\captionsetup{hypcap=true}
\label{fig:teaser}
\vspace{-1mm}
\end{strip}

\begin{abstract}
Image-to-3D reconstruction can disrupt image watermarks by regenerating object appearance and rendering it from new viewpoints.
We present \method, an image watermarking framework for blind recovery across viewpoint changes and unseen 3D reconstructors.
\method embeds a 32-bit message using content-attached RGB carriers computed from frozen DINO and VGG features by a lightweight network shared between its encoder and decoder.
A matched decoder re-estimates the carriers from a single render, requiring no source image, 3D asset, camera pose, or reconstructor identity.
The encoder and decoder are trained jointly, while embedding is feed-forward and requires no per-image optimization or reconstruction model at deployment.
At $35\,\mathrm{dB}$ foreground PSNR on held-out objects, training with 2D augmentations alone achieves $77.7\%$ mean bit accuracy across eight unseen reconstructors.
Training through one frozen reconstructor, TripoSR, yields $86.2\%$ on TripoSR and $83.7\%$ mean bit accuracy across seven unseen models spanning NERFs, textured meshes, and GSplats, outperforming
SOTA image-watermarking baselines, which remain near chance.
\vspace{-1mm}
\end{abstract}

\vspace{-1mm}
\section{Introduction}
\label{sec:intro}

\begingroup
\renewcommand{\thefootnote}{}
\footnotetext[1]{Project page: \url{https://github.com/Akou-v1/WIS3R}}
\endgroup

\begingroup
\renewcommand{\thefootnote}{\fnsymbol{footnote}}
\footnotetext[2]{In accordance with the ERC Open Access mandate, the authors have made the Author Accepted Manuscript (AAM) publicly available under the Creative Commons Attribution (CC-BY 4.0) license.}
\endgroup

Invisible image watermarking supports provenance by embedding hidden messages in published images. Learned encoder--decoder methods recover these messages after compression, geometric distortions, and generative image editing~\cite{zhu2018hidden,tancik2020stegastamp,bui2023trustmark,sander2025wam,lu2025vine}.
Recent image-to-3D (\iithreed) models introduce a different route for image reuse: a single object image can become a reusable 3D asset within seconds~\cite{hong2024lrm,tochilkin2024triposr,boss2024sf3d,zou2024tgs,szymanowicz2024splatter,xiang2025trellis}.
The asset or its renders can then be redistributed, preserving the object's appearance while potentially losing its provenance.
In our evaluation, released image-watermarking baselines fall to near-chance bit accuracy after reconstruction (\cref{tab:main}). 
One response is to disrupt reconstruction, as in Geometry Cloak~\cite{song2024geometrycloak} and concurrent work Flatland~\cite{flatland}, which protect source images by altering the reconstructed structure.
However, an owner may permit a character image to become a game asset, or a product photograph to become a 3D storefront model.
In such workflows, reconstruction is the intended outcome: the goal is to retain provenance while keeping the resulting asset useful.

Preserving a hidden message through this pipeline poses several challenges.
First, reconstruction regenerates appearance through learned representations and can suppress  small watermark perturbations.
Second, viewpoint changes alter projection, scale, and visibility, making detection patterns tied to source-image coordinates unreliable.
Third, the reconstructor may be unknown, so recovery must transfer across architectures and 3D representations without model-specific adaptation.
The watermark must meet these requirements while keeping the source image visually faithful.

We present \method, an image watermarking framework for provenance through \iithreed reconstruction (\cref{fig:teaser}).
To the best of our knowledge, it is the first to combine source-image embedding, multi-bit recovery from a single render after single-image 3D reconstruction, and zero-shot transfer to unseen reconstructors.
Recovery is \emph{blind}: the decoder does not require the source image, reconstructed asset, camera pose, or reconstructor identity.
To address regeneration and viewpoint changes, the encoder and decoder share a learned network that maps frozen DINO and VGG features to message-independent RGB fields, one per bit.
The encoder combines these \emph{content-attached carriers} according to the message. 
Conditioning their construction on object appearance allows the decoder to re-estimate corresponding carriers from a new render and recover the bits by correlation with color evidence, without explicit
source-to-render registration.
Embedding is feed-forward, with no per-image optimization or access to a reconstructor at publishing time.

Our main finding is \emph{zero-shot transfer}.
At $35\,\mathrm{dB}$ foreground PSNR, on held-out objects, our method trained using only 2D augmentations, achieves $77.7\%$ mean bit accuracy across eight unseen reconstructors spanning
NERFs, textured meshes, and Gsplats.
Training the same architecture through one frozen reconstructor, TripoSR, \method yields $86.2\%$ accuracy on TripoSR and $83.7\%$ mean accuracy across the seven unseen models, compared with $52.7\%$ for the strongest released baseline on those seven models.
Both variants outperform all evaluated released baselines on every tested reconstructor. 
To summarize, our contributions are:
\begin{itemize}[leftmargin=1.2em,itemsep=1pt,topsep=2pt]
  \item \textbf{Blind provenance through 3D reconstruction.} We introduce a learned appearance-conditioned RGB carrier construction and matched blind readout for recovering source-image messages from individual renders, without source-to-render registration or per-image optimization.
  \item \textbf{Zero-shot transfer across reconstructors.} We demonstrate recovery across 8 \iithreed pipelines using only 2D augmentations, and improved recovery on 7 unseen reconstructors when training through TripoSR alone, outperforming released watermarkers across all tested models.
\end{itemize}

\section{Related Work}
\label{sec:related}

\paragraph{Image watermarking.}
Learned encoder--decoder methods jointly optimize watermark embedding and recovery under simulated distortions~\cite{zhu2018hidden,tancik2020stegastamp,jia2021mbrs,ma2022cin,sander2025wam,bui2023trustmark}.
Image regeneration~\cite{zhao2023removable} and single-image novel-view synthesis~\cite{shamshad2026raven} expose vulnerabilities under generative transformations. 
Generation-time methods embed watermarks by controlling the initial sampling noise~\cite{wen2023treering} or fine-tuning the generative decoder~\cite{fernandez2023stablesignature}.
VINE~\cite{lu2025vine} and WaTeRFlow~\cite{jeong2026waterflow} target generative image editing and image-to-video generation.
\method instead targets reconstruction of a reusable 3D asset followed by rendering. 
Its decoder re-estimates content-conditioned RGB carriers from a single rendered view, enabling the same encoder--decoder pair to transfer to unseen reconstructors.

\paragraph{Carriers and content-based synchronization.}
Spread-spectrum watermarking motivates our correlation-based readout and host-interference compensation~\cite{cox1997spread,malvar2003iss}.
Feature-based synchronization attaches watermarks to image features~\cite{kutter1999second,bas2002geometrically}, while SSyncOA normalizes object location, orientation, and scale to resist crop-paste attacks~\cite{zhao2024ssyncoa}.
SpreadMark combines dense pseudo-random codewords with a learned chip representation and matched-filter readout~\cite{song2026spreadmark}.
Fernandez \etal optimize each marked image in a self-supervised feature space~\cite{fernandez2022sslwm}.
Our encoder and decoder share a learned mapping from host-image features to per-bit spatial RGB carriers, applied independently before and after \iithreed reconstruction. 
This supports blind readout without source-to-render registration and feed-forward embedding without per-image optimization.

\begin{figure*}[ht!]
  \centering
  \includegraphics[width=\textwidth]{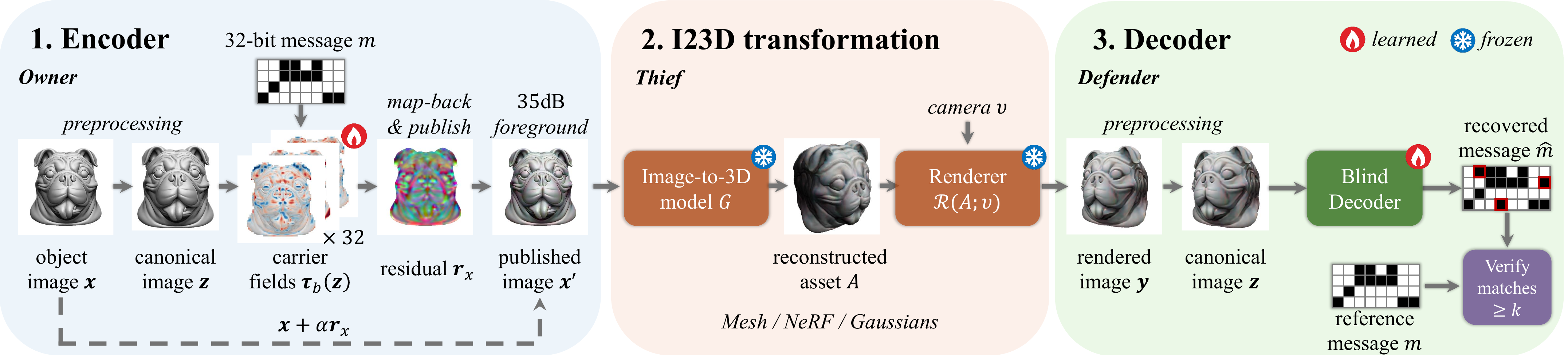}
  \vspace{-7mm}
  \caption{\textbf{\method pipeline.} The owner embeds a message in the foreground of $x$ and publishes $x'$ at a target foreground PSNR (35\,dB). 
  An unchanged reconstructor and renderer produce $y=\mathcal R(G(x');v)$. The decoder reads one render, using a supplied or estimated foreground mask, without knowing the source image,
  asset, reconstructor, or camera. The reference message is used only to check the decoded bits.
  }
  \vspace{-3mm}
  \label{fig:overview}
\end{figure*}

\paragraph{Watermarking 3D assets.}
Existing methods watermark radiance fields~\cite{luo2023copyrnerf,li2023steganerf,jang2024waterf,song2024nerfprotector},
Gaussian representations~\cite{huang2024gaussianmarker,chen2025guardsplat,li2025gsmarker,li2026w2m},
or meshes whose renders reveal a message~\cite{yoo2022deep3d2d}.
GaussianSeal and InstantSplamp integrate payloads into 3D generators~\cite{li2025gaussianseal,li2025instantsplamp}.
These approaches require access to a supported 3D representation or control over its creation. 
Generalization across assets in that setting does not establish whether a source-image watermark survives an unknown reconstructor.
\method instead embeds in the source image and reads a single render, without access to the intervening asset or reconstruction model at deployment.

\paragraph{Protecting images from 3D reconstruction.}
MantleMark is closest in protecting source images: it projects a 3D frequency-domain watermark into posed multi-view images and detects its presence in subsequent radiance-field renders using pose-dependent templates~\cite{luo2026mantlemark}.
\method instead embeds a multi-bit message in one image and recovers it from one render without its camera pose.
Geometry Cloak encodes customized messages as visible reconstructed geometry~\cite{song2024geometrycloak}, while concurrent work Flatland targets reconstruction disruption~\cite{flatland}.
These two approaches deter reconstruction by altering the generated structure.
\method seeks provenance through successful reconstruction: a hidden multi-bit message must survive appearance regeneration and viewpoint changes across unseen reconstructors. 
This supports attribution of redistributed assets and renders while aiming to retain useful object appearance.
\section{Method}
\label{sec:method}

\paragraph{Overview.}
\method uses a trained encoder to embed a binary message in an object image and a decoder to recover it from a single rendered view after image-to-3D (I23D) reconstruction. 
An attacker may convert the watermarked image into a reusable 3D asset and redistribute it or its renders (\cref{fig:teaser}). 
The decoder is \emph{blind}: it does not require the source image, the reconstructed asset, the camera pose, or the identity of the reconstruction model.

\paragraph{Inputs and outputs.}
The encoder takes an RGB object image $\vx\in[0,1]^{H\times W\times3}$, a foreground mask $\mM$, and a binary message $\vm\in\{0,1\}^{n}$, with $n=32$. 
We use the mask to concentrate the watermark on the object: I23D pipelines commonly remove or replace the background, potentially discarding watermark energy embedded there before reconstruction. 
The mask may be supplied by the owner or estimated by foreground segmentation \cite{gatis2020rembg}.
The encoder outputs a marked image $\vx'$ whose residual is confined to the foreground. 
An \emph{unmodified} I23D model $G$ reconstructs an asset $A$, which a renderer $\mathcal R$ converts into an image $\vy$ from viewpoint $v$:
\begin{equation}
    A = G(\vx'), \qquad \vy = \mathcal R(A;v).
    \label{eq:reconstruction_channel}
\end{equation}
The render may also be captured from an application using the asset. 
Given one image $\vy$ and its foreground $\mM_y$, supplied or estimated from $\vy$, the decoder estimates the bits $\hat{\vm}$. 
The owner's message $\vm$ is used only to verify this output.

\paragraph{Pipeline.}
\method introduces a learned watermark construction and matched blind readout for recovery through unseen I23D models. 
Inspired by spread-spectrum embedding~\cite{cox1997spread} and feature-based watermark  synchronization~\cite{bas2002geometrically}, we use a carrier-generation network (\Cref{sec:method:carrier-generation-network}) whose weights are shared between the encoder and decoder. 
This network maps image features to message-independent RGB fields, which we call \emph{content-attached carriers}. 
Each carrier specifies the spatial pattern and local color direction used to embed one message bit (\Cref{fig:carriers}). 
Conditioning the carriers on image content allows the encoder to construct them from the source image and the decoder to estimate the corresponding carriers from a received render. 
The encoder combines the carriers using message-dependent coefficients to embed the watermark (\Cref{sec:method:message-embedding}). 
The decoder correlates its estimated carriers with color evidence extracted from the render and decodes the message bits from the resulting scores (\Cref{sec:method:message-reading}). 
\Cref{fig:overview} shows the overall pipeline with its components, which we explain in detail in the following sections. 
We first discuss the components at deployment time, then we discuss the training stage in \Cref{sec:method:training}.

\subsection{Carrier-generation network}
\label{sec:method:carrier-generation-network}

The encoder and decoder share a carrier-generation network that produces $n$ content-conditioned RGB fields, one per message bit, without taking the message as input. 
For embedding, it processes the source image $\vx$ and mask $\mM$; for reading, it processes the received render $\vy$ and mask $\mM_y$.

\paragraph{Preprocessing.}
We crop a square around the foreground, composite it onto a gray background, and resize the image and mask to $512\times512$.
This standardizes object framing and resolution while limiting dependence on background appearance.
We denote the resulting canonical image by $\vz$.

\paragraph{Feature representation.}
We combine frozen DINO~\cite{caron2021dino} and VGG16~\cite{simonyan2015vgg} features to capture complementary cues about object structure and local appearance.
Let $F_0(\vz)$ denote their concatenation after resampling them to a common $64\times64$ grid.
To reduce left--right orientation bias, we average features from the image and its horizontally flipped copy after restoring spatial alignment:
\begin{equation}
  F(\vz)=\operatorname{std}\!\left(
  \tfrac12\big[
  F_0(\vz)+
  \operatorname{flip}\!\left(F_0(\operatorname{flip}(\vz))\right)
  \big]\right),
  \label{eq:shared_features}
\end{equation}
where $\operatorname{flip}$ horizontally reflects an image or feature grid, and $\operatorname{std}$ denotes feature standardization.
The DINO\ and VGG16\ backbone configurations details are provided in the supplement.

\paragraph{Carrier fields.}
A pointwise MLP $f_\theta$ with learned parameters $\vtheta$ maps each grid cell's features to $n$ RGB vectors, producing a field $\vt_b(\vz)=f_{\theta,b}(F(\vz))$ for each bit $b$.
To limit broad color shifts in the marked image, we suppress locally smooth components before normalizing each field:
\begin{equation}
  \vtau_b(\vz)=
  \mathcal H[\vt_b(\vz)] / 
       \operatorname{RMS}\!\left(\mathcal H[\vt_b(\vz)]\right),
  \label{eq:carrier}
\end{equation}
where $\mathcal H$ subtracts a foreground-restricted Gaussian local mean, and $\operatorname{RMS}$ computes the root mean square over foreground cells and RGB channels.
Both operations use the foreground mask associated with $\vz$.
Unit foreground RMS puts all carriers on a common scale before the encoder applies its embedding coefficients.
The resulting carrier tensor is $\vtau(\vz)\in\mathbb R^{n\times64\times64\times3}$.
The MLP architecture is specified in the supplement.

\begin{figure}[t!]
  \centering  
  \includegraphics[width=\columnwidth]{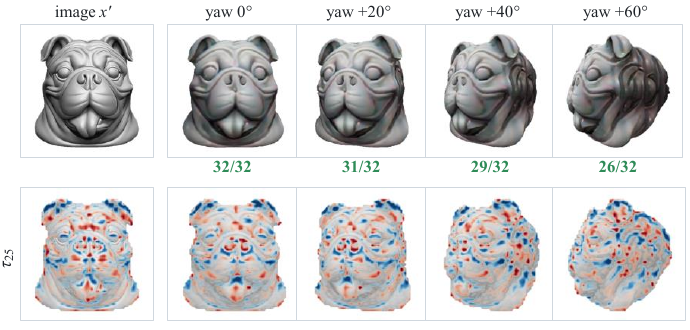}
  \vspace{-7mm}
  \caption{\textbf{Content-attached carriers across views.} The shared network generates carrier fields from the marked source image and renders from different viewpoints.
  Colors show the sign and magnitude of a carrier's response, illustrating approximate consistency of its appearance-conditioned structure across views.}
  \vspace{-3mm}
  \label{fig:carriers}
\end{figure}

During decoding, the same MLP recomputes the carriers from the received render. 
Because each carrier is computed from the local appearance of the object rather than from pixel positions, it tends to move with the surface; training through an I23D model makes the carriers agree across views. 
In \cref{fig:carriers}, the patches of one carrier stay on the same spots of the object as the asset turns from $0\dg$ to $60\dg$, and the decoder reads 26--32 of the 32 bits. 

\subsection{Message embedding}
\label{sec:method:message-embedding}

Given a canonical source image $\vz$ and its foreground mask $\mM_z$, the encoder combines the source image carriers using message-dependent coefficients with an embedding strength, and maps the resulting residuals back to the original image.

\paragraph{Host-interference compensation.}
We first convert the message to signs $\vs=2\vm-\vone\in\{-1,+1\}^{n}$.
The clean image can produce nonzero reader scores when its natural color variations correlate with the carriers estimated by the reader.
To estimate this host-induced bias, we evaluate the reader (\cref{sec:method:message-reading}) on $\vz$ before embedding.
Let $\ell_b(\vz)$ denote its scalar score for bit $b$. The embedding coefficients are

\begin{equation}
  \hat h_b=\ell_b(\vz),
  \qquad
  c_b=s_b-\lambda\hat h_b,
  \label{eq:embedding_coefficients}
\end{equation}
where the learned $\lambda$ controls the compensation strength.
During training, the clean-image scores are treated as constants. This construction follows the principle of host-interference compensation in spread-spectrum watermarking~\cite{malvar2003iss}. The encoder then forms the canonical image-resolution residual:
\begin{equation}
  \vr=\mM_z\odot\mathcal U\!\left(
      \sum_{b=1}^{n}c_b\vtau_b(\vz)
    \right),
  \label{eq:mark}
\end{equation}
where $\mathcal U$ upsamples the combined RGB field from $64\times64$ to $512\times512$, and $\odot$ applies the foreground mask to each color channel.
Masking confines the watermark to the object.

\paragraph{Publication.}
We map the residual back to the source image, $\vr_x=\mM\odot\mathcal B_x(\vr)$, where $\mathcal B_x$ resizes it into the source crop, and remove its mean over the foreground. 
The published image is
\begin{equation}
  \vx'=Q_8\!\left(\operatorname{clip}_{[0,1]}(\vx+\alpha\vr_x)\right),
  \label{eq:published_image}
\end{equation}
where $Q_8$ rounds to 8 bits. We choose the strength $\alpha$ for a foreground PSNR of $35\,\mathrm{dB}$, \ie the PSNR between $\vx$ and the clipped image before rounding, computed over object pixels only. 
The background and framing of $\vx$ are unchanged.

\subsection{Message reading}
\label{sec:method:message-reading}

Given a received render $\vy$ and its foreground mask $\mM_y$, the decoder applies the preprocessing of \cref{sec:method:carrier-generation-network} to obtain a canonical image $\vz$ and mask $\mM_z$; this is derived from the render itself, so no alignment to the source image is needed.
The raw reader produces bit scores $\vell(\vz)\in\mathbb R^n$ and a carrier-overlap matrix $\mK(\vz)\in\mathbb R^{n\times n}$, which are used for joint decoding below.
The same raw readout is used for clean-host compensation and reading marked source images; $\vz$ is the canonical image currently being processed.
The shared carrier-generation network recomputes $\vtau_b(\vz)$ using \cref{eq:carrier}.

\paragraph{Color filtering.} 
Since the watermark is embedded as RGB perturbations, the reader extracts color on the carrier grid:
\begin{equation}
  \vu=\mathcal H[\mathcal D(\vz)],
  \label{eq:reader_color_detail}
\end{equation}
where $\mathcal D$ resamples the image to $64\times64$ and $\mathcal H$ is the foreground-restricted high-pass operator defined in \cref{sec:method:carrier-generation-network}.
To reduce interference from natural texture, we weight color directions according to their local variability.
Let $\mSigma_q\in\mathbb R^{3\times3}$ denote the covariance of $\vu$ in a foreground-restricted Gaussian neighborhood of cell $q$, and let $\bar\sigma^2$ denote the foreground mean color variance.
We compute
\begin{equation}
    \mW_q
      =\bar\sigma^2
        \left(
          \mSigma_q+\epsilon\bar\sigma^2\mI_3
        \right)^{-1}, \qquad
    \tilde{\vu}_q=\mW_q\vu_q,
  \label{eq:whiten}
\end{equation}
where $\mI_3$ is the identity matrix and $\epsilon>0$ is diagonal loading term. 
This inverse-covariance weighting dampens directions with large local color variance, where host interference is stronger.
Covariance-estimation details and filter parameters are provided in the supplement.

\paragraph{Color evidence and scores.}
An extractor U-Net predicts an RGB watermark-evidence field $\hat{\vdelta}(\vz)\in\mathbb R^{64\times64\times3}$.
This learned branch complements the covariance-weighted color detail by extracting spatial patterns useful for message recovery.
Learned scalars $a_1$ and $a_2$ calibrate the contributions of the two evidence sources:
\begin{equation}
  \vv_q=a_1\tilde{\vu}_q+a_2\hat{\vdelta}_q.
  \label{eq:reader_evidence}
\end{equation}
Different object regions can provide different amounts of useful watermark evidence.
A $1\times1$ convolution of the shared features $F(\vz)$ therefore predicts a scalar spatial weight $\rho_q$ for each cell, allowing the reader to modulate its contribution to message recovery.
The raw score for bit $b$ is computed as:
\begin{equation}
  \ell_b(\vz)=
    \frac{1}{N_{\mathrm{fg}}}
    \sum_{q\in\mathrm{fg}}
      \rho_q
      \left\langle
        \vtau_{b,q},\vv_q
      \right\rangle,
  \label{eq:score}
\end{equation}
where the sum runs over the $N_{\mathrm{fg}}$ foreground cells on the carrier grid.
Here, $\vtau_{b,q}$ is the RGB vector of the carrier $b$ at cell $q$; dependence on $\vz$ is omitted from cell-indexed quantities for brevity.
Averaging over foreground cells normalizes the score aggregation over their count.

\paragraph{Carrier overlap.}
The carriers need not be orthogonal on a given image.
Consequently, evidence associated with one embedded bit can contribute to other bits' scores.
The reader estimates this coupling in the covariance-weighted correlation branch through
\begin{equation}
  K^{\mathrm{raw}}_{bd}(\vz)=
    \frac{1}{N_{\mathrm{fg}}}
    \sum_{q\in\mathrm{fg}}
      \rho_q
      \left\langle
        \vtau_{b,q},
        \mW_q\vtau_{d,q}
      \right\rangle.
  \label{eq:raw_carrier_overlap}
\end{equation}
We normalize the resulting matrix by its mean diagonal:
\begin{equation}  
    \kappa(\vz)
      =\frac{1}{n}
        \operatorname{tr}\!\left(
          \mK^{\mathrm{raw}}(\vz)
        \right), \qquad
    \mK(\vz)
      =\frac{\mK^{\mathrm{raw}}(\vz)}{\kappa(\vz)}.
  \label{eq:crosstalk}
\end{equation}
Thus, $\mK$ has unit mean diagonal, placing the overlap estimates on a common scale across images.
Its off-diagonal entries quantify pairwise carrier overlap under the reader's color and spatial weighting.
We use $\mK$ only as an approximate correction for the coupling between bits.

\paragraph{Joint decoding.}
At inference, the decoder uses the overlap matrix to jointly estimate all message signs by solving
\begin{equation}
  \left(\mK(\vz)+\mI_n\right)\hat{\vs}
    =\vell(\vz),
  \label{eq:joint}
\end{equation}
where $\hat{\vs}\in\mathbb R^n$ contains the estimated signs. 
The identity term regularizes the approximate overlap correction.
Since $\mK$ has unit mean diagonal, this regularization uses a common reference scale across images.
The decoded bits are
\begin{equation}
  \hat m_b=\mathbbm{1}[\hat s_b>0].
  \label{eq:decoded_bits}
\end{equation}
Decoding uses a single render, network evaluations, and an $n\times n$ linear solve, without source-image access, camera-pose estimation, or reconstruction-model adaptation.

\begin{table*}[ht!]
\centering
\caption{\textbf{Image-watermark recovery after I23D reconstruction.} Bit accuracy on the Objaverse test split; each render is read independently.
Neural methods use a $35\,\mathrm{dB}$ foreground PSNR target.
$\star$: the sole training reconstructor for \method.
The final columns average all eight models and the seven excluding TripoSR.
All eight are unseen during \method-2D training.
$\dagger$: DWT-DCT uses its native strength and is not distortion matched.
Both variants exceed every released baseline on every reconstructor.}
\vspace{-3mm}
\label{tab:main}
\setlength{\tabcolsep}{3pt}
\resizebox{\textwidth}{!}{%
\begin{tabular}{lccccccccccc}
\toprule
& & \multicolumn{8}{c}{Bit accuracy $\uparrow$}
& \multicolumn{2}{c}{Mean $\uparrow$} \\
\cmidrule(lr){3-10}\cmidrule(lr){11-12}
Method & Bits & TripoSR$^\star$ & Real3D & OpenLRM & VFusion3D & SF3D
& TGS & Splatter & TRELLIS & All 8 & Unseen \\
\midrule
VINE \cite{lu2025vine} & 100 & 49.9\% & 50.1\% & 50.3\% & 49.7\% & 50.1\% & 49.9\% & 50.1\% & 49.0\% & 49.9\% & 49.9\% \\
TrustMark \cite{bui2023trustmark} & 100 & 53.2\% & 53.5\% & 51.3\% & 51.8\% & 53.4\% & 52.2\% & 53.2\% & 49.6\% & 52.3\% & 52.1\% \\
HiDDeN \cite{zhu2018hidden} & 48 & 51.3\% & 51.7\% & 52.2\% & 52.5\% & 54.5\% & 51.4\% & 51.0\% & 51.0\% & 51.9\% & 52.0\% \\
MBRS \cite{jia2021mbrs} & 256 & 50.1\% & 50.0\% & 49.9\% & 49.9\% & 50.5\% & 50.1\% & 50.0\% & 50.0\% & 50.1\% & 50.1\% \\
CIN \cite{ma2022cin} & 30 & 50.7\% & 52.0\% & 50.4\% & 51.3\% & 51.9\% & 50.3\% & 50.6\% & 49.3\% & 50.8\% & 50.8\% \\
WAM \cite{sander2025wam} & 32 & 53.1\% & 52.7\% & 53.2\% & 52.5\% & 54.6\% & 51.5\% & 53.7\% & 50.9\% & 52.8\% & 52.7\% \\
VideoSeal 1.0 \cite{fernandez2024videoseal} & 256 & 50.7\% & 51.4\% & 50.9\% & 51.8\% & 53.9\% & 52.3\% & 52.3\% & 50.2\% & 51.7\% & 51.8\% \\
PixelSeal \cite{soucek2025pixelseal} & 256 & 50.5\% & 50.4\% & 50.4\% & 51.7\% & 53.3\% & 50.4\% & 50.5\% & 50.2\% & 50.9\% & 51.0\% \\
DWT-DCT$^\dagger$ \cite{invisiblewatermark} & 32 & 49.0\% & 49.0\% & 48.8\% & 49.0\% & 49.0\% & 49.0\% & 49.0\% & 48.4\% & 48.9\% & 48.9\% \\
\hline
\method-2D & 32 & \cellcolor{custom_green!30}{78.2\%} & \cellcolor{custom_green!30}{82.1\%} & \cellcolor{custom_green!30}{80.8\%} & \cellcolor{custom_green!30}{79.7\%} & \cellcolor{custom_green!30}{85.6\%} & \cellcolor{custom_green!30}{75.0\%} & \cellcolor{custom_green!30}{77.4\%} & \cellcolor{custom_green!30}{63.1\%} & \cellcolor{custom_green!30}{77.7\%} & \cellcolor{custom_green!30}{77.7\%} \\
\method & 32 & \cellcolor{custom_green!70}{86.2\%} & \cellcolor{custom_green!70}{88.1\%} & \cellcolor{custom_green!70}{88.2\%} & \cellcolor{custom_green!70}{86.3\%} & \cellcolor{custom_green!70}{90.4\%} & \cellcolor{custom_green!70}{81.0\%} & \cellcolor{custom_green!70}{84.4\%} & \cellcolor{custom_green!70}{67.8\%} & \cellcolor{custom_green!70}{84.0\%} & \cellcolor{custom_green!70}{83.7\%} \\
\bottomrule
\end{tabular}%
}
\vspace{-2mm}
\end{table*}

\subsection{Training}
\label{sec:method:training}

We jointly train the carrier construction, embedding coefficients, and raw reader for message recovery under image transformations or reconstruction and rendering.

\paragraph{Training examples and channels.}
Each example consists of a source image $\vx$, foreground mask $\mM$, and randomly sampled message $\vm\in\{0,1\}^{n}$.
The encoder produces the published image $\vx'$ using \cref{sec:method:message-embedding}.
We generate $V=4$ channel outputs from the same marked image and message, exposing each training example to several transformations or views.
Geometric transformations are also applied to the foreground mask; rendered outputs use their corresponding foreground masks.
All outputs are canonicalized before reading. We consider two channel configurations with the same learned components and objective:
\begin{itemize}
  \item \textbf{2D-only training.}
  Each of the $V$ copies receives one sampled transformation:
  JPEG compression, blur, in-plane rotation, cropping, additive noise, resizing, horizontal flipping, color jitter, or identity.
  No 3D reconstructor is used in this variant. We refer to this variant as \textbf{``\method-2D''}.

  \item \textbf{Reconstruction-aware training.}
  We reconstruct a colored mesh from $\vx'$ using frozen TripoSR~\cite{tochilkin2024triposr} and render it with nvdiffrast~\cite{laine2020nvdiffrast}.
  The $V$ cameras are sampled across the four yaw quadrants, providing view diversity from a single reconstructed asset.
  TripoSR is the only reconstructor used during training.
  Gradients propagate through the predicted vertex colors and their rendering, while the extracted mesh geometry is held constant in the backward pass.
  This retains an appearance-based training signal without differentiating through the discrete mesh extraction.
  The forward pass reconstructs the current marked input, including its effect on geometry. 
  This full variant (\textbf{``\method''}) is the best performing one in our experiments.
\end{itemize}

Augmentation distributions, camera sampling, and mesh-extraction settings are detailed in the supplement.

\paragraph{Recovery objective.}
Let $\vz_0$ denote the canonicalized published image $\vx'$, and let $\vz_j$ denote the canonicalized output of channel copy or view $j$.
For each source example, we apply binary cross-entropy with logits to the raw reader scores:
\begin{equation}
    \mathcal L_{\mathrm{msg}}={}
      \frac{1}{V}\sum_{j=1}^{V}
      \operatorname{CE}
        \!\left(\vell(\vz_j),\vm\right)
      +\beta\,
      \operatorname{CE}
        \!\left(\vell(\vz_0),\vm\right).
  \label{eq:message_loss}
\end{equation}
where $\operatorname{CE}$ averages over the $n$ bits and $\beta=0.5$.
The channel-output term encourages survival through the training channel, while the source-image term retains readability before reconstruction or other transformations.
We note that the joint solve in \cref{eq:joint} is used only at inference and is not applied inside the recovery loss.

\paragraph{Optimization.}
We optimize the carrier MLP, evidence extractor, spatial-weight predictor, mixing scalars $a_1,a_2$, and compensation strength $\lambda$ jointly using Adam on the minibatch mean of the message loss. 
WIS3R-2D and WIS3R are trained independently from scratch with identical settings. 
DINO, VGG16 and TripoSR retain fixed weights. 
Architectural details, minibatch sizes, training duration, and the learning-rate schedule are provided in the supplement.

\section{Experiments}
\label{sec:experiments}

\begin{figure*}[th!]
  \centering
  \includegraphics[width=\textwidth]{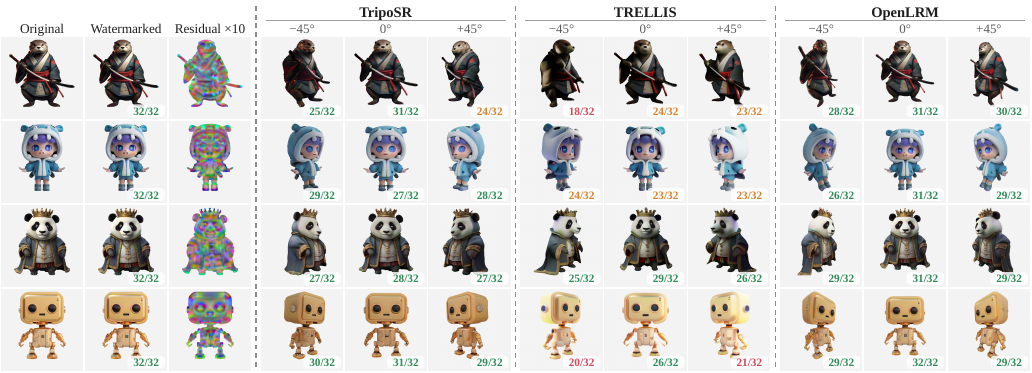}
  \vspace{-6mm}
  \caption{\textbf{Representative qualitative results.} From left to right: source images, marked versions ($35\,\mathrm{dB}$ foreground PSNR), and residuals ($\times10$), followed by renders at yaws $-45^\circ$, $0^\circ$, and $+45^\circ$. 
  TripoSR is the sole training reconstructor; TRELLIS and OpenLRM are unseen. Counts show correctly decoded bits out of 32 from each image independently (green: $\geq25$, yellow: $\geq23$, red: $<23$).}
  \vspace{-3mm}
  \label{fig:qual}
\end{figure*}

We now discuss our training and evaluation protocol, along with comparisons, results, and ablations.
The goal of our evaluation is whether a watermark embedded in a single image remains readable after reconstruction and rendering, and whether this ability transfers to reconstructors excluded from training.

\subsection{Evaluation protocol}
\label{sec:setup}

\paragraph{Data and messages.}
We use Objaverse~\cite{deitke2023objaverse}, with $20K$ objects for training, $500$ for validation, $500$ for testing. 
Each object provides one $512\times512$ source image.
To test transfer beyond Objaverse, we evaluate on $250$ objects from Google Scanned Objects (GSO)~\cite{downs2022gso} and $250$ from Toys4K~\cite{stojanov2021toys4k}.
Our method variants embed a random $32$-bit message per test object, drawn from a pool excluded from training. 

\paragraph{Reconstruction and viewpoints.}
We evaluate eight released I23D pipelines: TripoSR~\cite{tochilkin2024triposr}, Real3D~\cite{jiang2024real3d}, OpenLRM-1.1~\cite{he2023openlrm}, VFusion3D~\cite{han2024vfusion3d}, SF3D~\cite{boss2024sf3d}, TGS~\cite{zou2024tgs}, Splatter Image~\cite{szymanowicz2024splatter}, and
TRELLIS~\cite{xiang2025trellis}.
Their pipelines span radiance fields, textured meshes, and Gaussian splats; TRELLIS additionally tests transfer to a flow-based generator.
We use their released weights and rendering pipelines without adapting them to the watermark. Only TripoSR is used to train \method; \method-2D uses none of these models.
Each reconstructed asset is rendered eight times, with yaw sampled uniformly from $[-45^\circ,45^\circ]$ relative to the source view and elevation from $[-15^\circ,15^\circ]$.
Every render is decoded independently, without combining evidence across views.
This primary range emphasizes views overlapping the conditioning view and tests retention of evidence embedded in the observed appearance.
Larger view changes introduce surfaces whose appearance is not specified by the source image; we evaluate those separately in \cref{sec:range}.

\paragraph{Distortion and baseline protocol.}
Following \cref{sec:method:message-embedding}, our default strength is calibrated to $35\,\mathrm{dB}$ foreground PSNR after clipping and before 8-bit rounding.
The metric concerns the object pixels, excluding the background from the distortion budget.
We compare with several popular baselines: VINE~\cite{lu2025vine}, TrustMark~\cite{bui2023trustmark}, HiDDeN~\cite{zhu2018hidden}, MBRS~\cite{jia2021mbrs}, CIN~\cite{ma2022cin}, WAM~\cite{sander2025wam}, VideoSeal 1.0~\cite{fernandez2024videoseal}, PixelSeal~\cite{soucek2025pixelseal}, and DWT-DCT~\cite{invisiblewatermark}. 
Released baselines retain the payload lengths (listed in \cref{tab:main}).
For the neural baselines, we use the released encoders and decoders with the same foreground-centered preprocessing, confine the residual to the object, and rescale it to the same target foreground PSNR.
DWT-DCT is retained at its native embedding setting and is identified separately because its distortion is not matched.
Decoder settings are fixed across reconstructors.

\paragraph{Metrics.}
We report bit accuracy and message-level true-positive rate (TPR), following W-Bench~\cite{lu2025vine}, which sets the detection thresholds from the binomial tail of a random message at false-positive rates of $0.1\%$ and $1\%$. 
For $32$-bit messages, $\mathrm{TPR}_{25}$ and $\mathrm{TPR}_{23}$ are the fractions of watermarked renders with at least $25$ and $23$ correctly decoded bits.
The owner message is used only to score the blind output.
Thresholds remain fixed and identical across reconstructors and have nominal false-match probabilities of $0.105\%$ and $1.003\%$ for an independent uniformly random candidate message.
The false-positive rates of randomly sampled keys are also measured in the supplement.
Model averages weight reconstructors equally.

\subsection{Comparisons and zero-shot transfer}

\paragraph{Released watermarkers.}
\Cref{tab:main} shows a large separation after reconstruction: the strongest released baseline, WAM, reaches $52.8\%$ mean bit accuracy, compared with $77.7\%$ for \method-2D and $84.0\%$ for \method.
These are gains of $24.9$ and $31.2$ percentage points, respectively.
The equal-payload WAM comparison also controls for the number of embedded bits: both WAM and our variants carry $32$ bits.
Before reconstruction, WAM reads $99.4\%$ of its embedded bits and \method reads $99.5\%$.
Their similar source-image readability, followed by sharply different post-reconstruction recovery, identifies the I23D pipeline as the main challenge in our setting.

\paragraph{Transfer across reconstructors.}
With one fixed encoder--decoder pair trained through TripoSR alone, \method reaches $83.7\%$ mean bit accuracy on the seven unseen models.
Real3D, OpenLRM, and SF3D reach $88.1\%$, $88.2\%$, and $90.4\%$, exceeding the $86.2\%$ achieved through the training reconstructor.
Transfer also extends to the harder flow-based TRELLIS pipeline, where accuracy is $67.8\%$.
Thus, recovery is maintained across reconstruction pipelines and representations without model-specific adaptation, although its strength varies across them. 
The same fixed decision thresholds give $\mathrm{TPR}_{25}=80.6\%$ and $\mathrm{TPR}_{23}=90.0\%$ through TripoSR, and means of $72.9\%$ and $82.5\%$ across the seven unseen models.
Before reconstruction, all reported \method source images meet both thresholds.

\paragraph{Fine-tuning through the same reconstructor.}
To distinguish the proposed construction from exposure to 3D training alone, we also fine-tune VINE, WAM, and HiDDeN through the same frozen TripoSR reconstruction and rendering pipeline. 
Checkpoint selection uses validation data. 
The comparison uses a $32$-bit evaluated message, with unused positions padded where required, the same foreground preprocessing, and the $35\,\mathrm{dB}$ target.
WAM is the strongest adapted baseline at $62.7\%$ (\cref{tab:finetune}), while \method reaches $86.2\%$: a $23.5$ point gain. 
At the stricter threshold, the gap further increases -- their detection rates are $5.9\%$ and $80.6\%$, respectively. Per-method recipes and learning rates are given in the supplement.

\begin{table}[t!]
\centering
\caption{\textbf{Comparison after TripoSR fine-tuning.}
VINE, WAM, and HiDDeN are adapted through the same reconstruction and differentiable-rendering pipeline used by \method, with $32$ evaluated message bits and a $35\,\mathrm{dB}$ foreground target.
TPRs use the fixed match-count thresholds defined above.}
\vspace{-2mm}
\label{tab:finetune}
\small
\setlength{\tabcolsep}{4pt}
\begin{tabular}{lccc}
\toprule
Method & Bit acc. & $\mathrm{TPR}_{25}$ & $\mathrm{TPR}_{23}$ \\
\midrule
VINE & 51.3\% & 1.6\% & 1.6\% \\
WAM & 62.7\% & 5.9\% & 21.5\% \\
HiDDeN & 51.1\% & 0.0\% & 1.6\% \\
\method & \cellcolor{custom_green!70}{86.2\%} & \cellcolor{custom_green!70}{80.6\%} & \cellcolor{custom_green!70}{90.0\%} \\
\bottomrule
\end{tabular}
\vspace{-2mm}
\end{table}

\subsection{Robustness and ablations}

\paragraph{Distortion, compression, and dataset shift.}
\Cref{tab:conditions} evaluates the same encoder and reader at reduced embedding strengths, under JPEG compression, and on new object datasets.
At $40\,\mathrm{dB}$, obtained by rescaling the residual without retraining, bit accuracy remains $80.7\%$ through TripoSR and $78.9\%$ across unseen models.
JPEG quality $75$ applied only to the published image leaves unseen-model accuracy at $83.4\%$.
Compressing both the source and the received render is harder: unseen-model accuracy is $80.4\%$, and $\mathrm{TPR}_{25}$ decreases from $72.9\%$ to $64.8\%$ in this sweep.
On GSO and Toys4K, unseen-model accuracies reach $86.0\%$ and $84.8\%$, respectively, showing recovery extends beyond Objaverse.

\begin{table}[t!]
\centering
\caption{\textbf{Transfer under additional conditions.}
Bit accuracy and $\mathrm{TPR}_{25}$ for TripoSR and the mean of the seven unseen reconstructors. The default is Objaverse at $35\,\mathrm{dB}$.
JPEG uses quality $75$, applied before reconstruction (source) after rendering. GSO and Toys4K use the default strength.}
\vspace{-2mm}
\label{tab:conditions}
\setlength{\tabcolsep}{3pt}
\resizebox{\columnwidth}{!}{%
\begin{tabular}{lcccc}
\toprule
& \multicolumn{2}{c}{TripoSR} & \multicolumn{2}{c}{Unseen seven} \\
\cmidrule(lr){2-3}\cmidrule(lr){4-5}
Condition & Bit acc. & $\mathrm{TPR}_{25}$ & Bit acc. & $\mathrm{TPR}_{25}$ \\
\midrule
35\,dB & 86.2\% & 80.6\% & 83.7\% & 72.9\% \\
38\,dB & 83.0\% & 69.1\% & 81.0\% & 65.2\% \\
40\,dB & 80.7\% & 61.6\% & 78.9\% & 59.0\% \\
JPEG: source & 85.9\% & 78.9\% & 83.4\% & 72.0\% \\
JPEG: source + render & 85.1\% & 77.1\% & 80.4\% & 64.8\% \\
GSO & 90.5\% & 92.4\% & 86.0\% & 80.3\% \\
Toys4K & 86.2\% & 84.1\% & 84.8\% & 76.4\% \\
\bottomrule
\end{tabular}%
}
\vspace{-2mm}
\end{table}

\paragraph{Viewpoint range.}
\label{sec:range}
Recovery is strongest near the conditioning view and remains above chance at larger yaw changes, but detection becomes somewhat weaker (\Cref{fig:qual}).
Across unseen models, bit accuracy is $83.7\%$ in the primary range, $68.8\%$ at $45^\circ$--$90^\circ$ absolute yaw, $63.5\%$ at $90^\circ$--$135^\circ$, and $67.5\%$ at $135^\circ$--$180^\circ$.
The rear-view increase shows that performance does not decrease monotonically with yaw.
These measurements show the strongest recovery around the source view, with partial retention at larger view changes (see supplement for more discussion and results).

\paragraph{Fidelity.}
Every method's residual is scaled by a single factor so that the foreground PSNR is exactly $35\,\mathrm{dB}$, and the background is unchanged. 
At this budget, \method's published images reach an SSIM of $0.990$ and an LPIPS of $0.045$ against the source on the object region.

\paragraph{Ablations.}
\Cref{tab:ablation} separates fixed-weight interventions from variants retrained with a changed construction or objective.
Content conditioning has the largest effect: replacing the image-dependent carriers with an image-independent set reduces accuracy from $86.2\%$ to $63.9\%$ ($22.3$ percentage points).
Clamping DINO or VGG features to their mean reduces accuracy to $72.4\%$ and $75.7\%$, supporting the use of complementary structural and appearance cues. 
Reconstruction-aware training increases accuracy over \method-2D from $78.2\%$ to $86.2\%$ on TripoSR, while the 2D-only model already transfers to every tested reconstructor.
Among the reader components, disabling the learned evidence extractor reduces accuracy to $82.5\%$ and disabling inverse-covariance weighting reduces it to $84.7\%$.
The smaller components have mixed effects in this setting: joint decoding improves bit accuracy but not the reported fixed-threshold detection rates.

\begin{table}[t]
\centering
\caption{\textbf{Ablations on TripoSR.}
Objaverse test objects in the primary view range, at the $35\,\mathrm{dB}$
target. Upper interventions retain the trained weights; lower variants
are retrained. Feature interventions set the indicated features to their
mean in the encoder and decoder.}
\vspace{-2mm}
\label{tab:ablation}
\setlength{\tabcolsep}{3pt}
\resizebox{\columnwidth}{!}{%
\begin{tabular}{lccc}
\toprule
Variant & Bit acc. & $\mathrm{TPR}_{25}$ & $\mathrm{TPR}_{23}$ \\
\midrule
Full \method & \cellcolor{custom_green!70}{86.2\%} & \cellcolor{custom_green!30}{80.6\%} & \cellcolor{custom_green!30}{90.0\%} \\
\midrule
\multicolumn{4}{l}{\emph{Fixed-weight interventions}} \\
DINO features set to mean & 72.4\% & 38.6\% & 57.1\% \\
VGG features set to mean & 75.7\% & 49.9\% & 69.1\% \\
No inverse-covariance weighting & 84.7\% & 77.6\% & 88.4\% \\
No joint decoding & 85.2\% & \cellcolor{custom_green!70}{80.7\%} & \cellcolor{custom_green!70}{90.5\%} \\
No spatial weighting & \cellcolor{custom_green!30}{85.9\%} & 79.9\% & 88.2\% \\
No evidence extractor & 82.5\% & 67.6\% & 81.9\% \\
No host compensation & \cellcolor{custom_green!30}{85.9\%} & 80.0\% & 89.5\% \\
\midrule
\multicolumn{4}{l}{\emph{Retrained variants}} \\
\method-2D & 78.2\% & 56.4\% & 69.0\% \\
Image-independent carriers & 63.9\% & 24.0\% & 32.0\% \\
\bottomrule
\end{tabular}%
}
\vspace{-2mm}
\end{table}

\section{Conclusion}
\label{sec:conclusion}
\label{sec:limits}
We introduced \method, an image watermarking method for blind message recovery after single-image 3D reconstruction. 
Its content-attached carriers are computed from object appearance and re-estimated from a rendered view of the reconstructed asset. 
We showed zero-shot recovery of source-image messages across multiple unseen reconstructors.

\paragraph{Limitations. }
At $35\,\mathrm{dB}$ the mostly chromatic mark is hard to see on textured objects but can leave faint tints on uniform surfaces. 
Lowering the distortion to $40\,\mathrm{dB}$ reduces unseen-model accuracy to $78.9\%$  (\cref{tab:conditions}), which remains far above other baselines at $35\,\mathrm{dB}$. \method targets views that overlap the published image, where the render shows the appearance the mark was written into. 
Beyond the primary range the reconstructor has to invent more of the visible surface, and accuracy over the unseen models falls to $63.5$--$68.8\%$ on the sides and rear (\cref{sec:range}). 
These views still read clearly above the baselines, which stay close to chance even within the primary range. 
Still, detection from side or rear renders is less reliable, thus wider view changes remain open.

\section*{Acknowledgements}
This project has received funding from the European Research Council (ERC) under the Horizon Research and Innovation Programme (Grant agreement No.~101124742).

{
    \small
    \bibliographystyle{ieeenat_fullname}
    \bibliography{main}
}

\clearpage
\setcounter{page}{1}
\maketitlesupplementary
\appendix

\section{Implementation details}
\label{sec:supp:impl}

\begin{figure*}[t!]
  \centering
  \includegraphics[width=\textwidth]{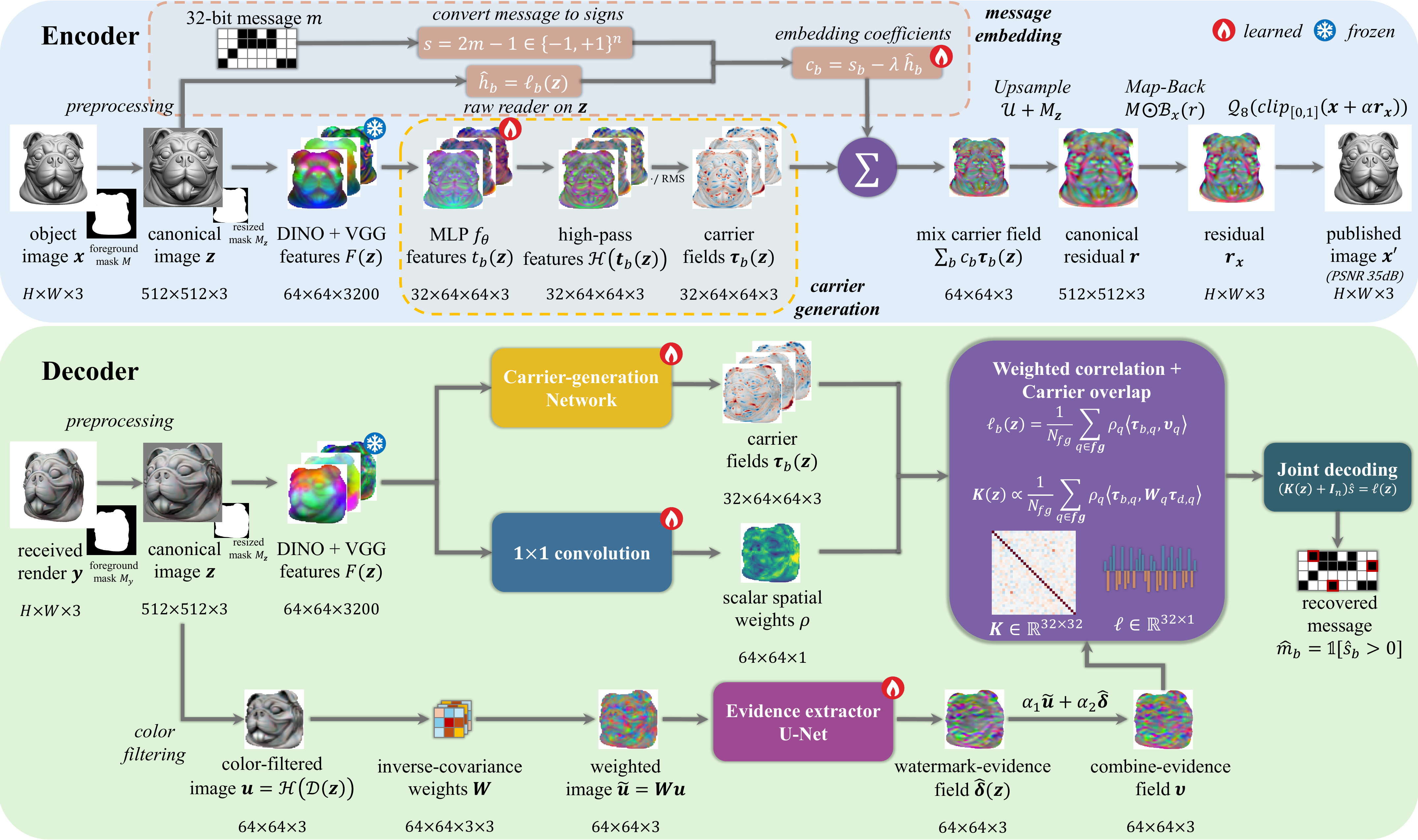}
  \vspace{-7mm}
\caption{\textbf{\method components and information flow.} Frozen backbones produce $64\times64\times3200$ features. 
The shared carrier MLP maps them to $32$ RGB fields, which the encoder combines using message signs and host-interference compensation. 
The residual is mapped back, masked, and scaled for publication. 
The reader independently recomputes carriers, combines filtered color detail with learned watermark evidence, and returns scores $\vell$ and a normalized overlap matrix $\mK$. 
Joint decoding is applied only at inference.}
  \vspace{-3mm}
  \label{fig:method}
\end{figure*}

This supplementary material provides implementation details and additional evaluation of \method.
\Cref{sec:supp:impl} describes the network architectures, preprocessing, carrier construction, embedding and reading procedures, and training setup.
\Cref{sec:supp:fpr} reports per-key false-positive measurements and analyzes pooled threshold calibration.
\Cref{sec:supp:baselines} specifies the evaluation protocol for released watermarkers and their reconstruction-aware fine-tuning.
\Cref{sec:supp:results} extends the evaluation to wider viewpoint ranges, reports source-image fidelity and reconstruction-pipeline configurations, and evaluates recovery with estimated foreground masks.

\subsection{Architecture details}
\label{sec:supp:architectures}
\Cref{fig:method} summarizes the components and information flow of \method. Unless stated otherwise, the two variants share the same architecture and hyperparameters; their training channels differ as described below. 

\Cref{tab:wis3r_encoder,tab:wis3r_decoder} specify embedding and reading, and \cref{tab:wis3r_shared,tab:wis3r_blocks} expand their shared components.
In these tables, $\mM_g$ denotes the foreground-cell mask derived from $\mM_z$, $N_{\mathrm{fg}}$ its foreground-cell count, and $\mF=F(\vz)$ the evaluated shared feature tensor. 
The vectors $\vmu_F$ and $\bm{\sigma}_F$ contain the fixed channel-standardization statistics and are broadcast across spatial locations.
The foreground mean $\mu_M$ is computed independently per channel. 
$\mathcal G_\sigma^{\mathrm{fg}}$ denotes Gaussian averaging normalized over foreground cells, with standard deviation $\sigma$ in grid cells. 
The high-pass operation applies the corresponding local-mean subtraction separately to every channel. Shapes omit the minibatch dimension.

\subsection{Canonical frame}
\label{sec:supp:frame}
Given an image and its foreground mask, we center a square crop on the mask's bounding box. Its side length is the longer box side divided by $0.85$. 
We composite the foreground onto gray (RGB value $0.5$) and resize the image to $512\times512$ using bicubic interpolation. The associated mask defines the foreground in this canonical frame.
The carrier and reader grids have $64\times64$ cells, each covering an $8\times8$-pixel region. 
A cell's color is the mean RGB value of its foreground pixels.

\subsection{Feature backbones}
\label{sec:supp:features}

\paragraph{DINO.}
We extract patch features from layers $4$, $8$, and $12$ of frozen DINO ViT-B/16~\cite{caron2021dino}, evaluated on the $512\times512$ canonical image. 
Each layer contributes $768$ channels, resampled bilinearly to the $64\times64$ carrier grid.

\paragraph{VGG16.} 
VGG16~\cite{simonyan2015vgg} runs on the canonical image resized to $256\times256$ (bicubic). 
We take \texttt{relu2\_2} ($128$ channels at $128^2$, average-pooled to $64^2$), \texttt{relu3\_3} ($256$ channels at $64^2$), and \texttt{relu4\_3} ($512$ channels at $32^2$, upsampled bilinearly to $64^2$).

\paragraph{Concatenation and standardization.}
Concatenation gives $F_0(\vz)\in\mathbb R^{64\times64\times3200}$, with $2304$ DINO and $896$ VGG channels. 
We average features from the image and its horizontal flip after restoring their spatial alignment, as in the main paper. 
Each channel is then standardized using a fixed mean and standard deviation. 
The same feature construction and statistics are used for embedding and reading; both backbones remain frozen.

\subsection{Carrier network and filters}
\label{sec:supp:carriers}
The pointwise MLP $f_\theta$ applies four $1\times1$ convolutions with widths $3200\to1024\to512\to256\to96$ ($3.96$\,M parameters). 
Each hidden layer is followed by per-cell LayerNorm and SiLU. The $96$ output channels are reshaped into $n=32$ RGB fields. 
The network receives no message.
$\mathcal H$ subtracts a foreground-restricted Gaussian local mean with standard deviation $6$ cells and sets background cells to zero. 
Each filtered carrier is normalized to unit RMS over foreground cells and RGB channels. 
Filtering limits slowly varying color shifts, while RMS normalization places the carriers on a common amplitude scale before message-dependent mixing.

\begin{figure}[t] 
  \centering \includegraphics[width=\columnwidth]{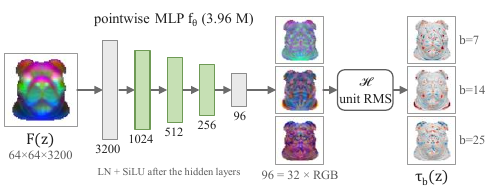}
  \vspace{-9mm}
  \caption{\textbf{Carrier network.} The pointwise MLP maps each cell's $3200$ feature channels to $32$ RGB vectors. 
  Foreground-restricted high-pass filtering and RMS normalization produce the message-independent carrier fields.}  
  \label{fig:supp:carrier}
  \vspace{-3mm}
\end{figure}

\subsection{Reader}
\label{sec:supp:reader}
\paragraph{Local color statistics.}
Let $\vu=\mathcal H[\mathcal D(\vz)]$ be the high-pass cell colors. 
We estimate local color variability using foreground-normalized Gaussian weights $w_{qr}$ with standard deviation $4$ cells:
\begin{equation}
  \mSigma_q=\sum_{r\in\mathrm{fg}}w_{qr}\vu_r\vu_r^\top,
  \qquad \sum_{r\in\mathrm{fg}}w_{qr}=1.
  \label{eq:supp:color_statistic}
\end{equation}
The implementation uses this uncentered second moment; no local-mean outer product is subtracted. The foreground mean per-channel second moment is
\begin{equation}
  \bar\sigma^2=
  \frac{1}{N_{\mathrm{fg}}}
  \sum_{q\in\mathrm{fg}}\frac{\operatorname{tr}(\mSigma_q)}{3}.
  \label{eq:supp:mean_variance}
\end{equation}
We compute
\begin{equation}
  \mW_q=\bar\sigma^2
    (\mSigma_q+\epsilon\bar\sigma^2\mI_3)^{-1},
  \qquad \tilde{\vu}_q=\mW_q\vu_q,
  \label{eq:supp:color_weighting}
\end{equation}
with $\epsilon=0.03$. 
This attenuates color directions with strong local variation, reducing interference from the object's natural appearance.

\paragraph{Evidence extractor.}
The evidence extractor receives $55$ channels per cell: foreground-pooled RGB colors ($3$), filtered color detail $\tilde{\vu}$ ($3$), a learned $1\times1$ projection of $F(\vz)$ from $3200$ to $48$ channels, and the foreground mask ($1$).
The U-Net uses channel widths $48$, $96$, and $192$, with the $192$-channel stage at $16\times16$ resolution, two average-pooling operations, bilinear upsampling, and skip connections (\cref{tab:wis3r_blocks}). 
Each processing stage contains two $3\times3$ convolution--GroupNorm--SiLU blocks, with $8$ GroupNorm groups and left--right symmetric convolution kernels. 
A final $1\times1$ convolution predicts the RGB evidence field $\hat{\vdelta}\in\mathbb R^{64\times64\times3}$.
This field is supervised through message recovery, without a separate residual-reconstruction loss.

\paragraph{Scores.}
The reader independently recomputes $\vtau_b(\vz)$ from the received canonical image using the shared carrier network. 
Learned scalars combine the two evidence sources as
\[
  \vv_q=a_1\tilde{\vu}_q+a_2\hat{\vdelta}_q.
\]
The raw score is their foreground-weighted correlation with the carrier:
\[
  \ell_b(\vz)=\frac{1}{N_{\mathrm{fg}}}
  \sum_{q\in\mathrm{fg}}\rho_q
  \left\langle\vtau_{b,q},\vv_q\right\rangle.
\]

\paragraph{Spatial weights.}
The scalar weight $\rho_q$ is obtained by applying softplus to a learned $1\times1$ convolution of $F(\vz)$ and dividing by the foreground mean. 
The resulting weights are nonnegative and average to $1$, so they redistribute spatial contributions without introducing a global scale factor.

\paragraph{Overlap matrix and joint decoding.}
The raw overlap matrix measures carrier coupling in the filtered-color branch:
\[
  K^{\mathrm{raw}}_{bd}(\vz)=
  \frac{1}{N_{\mathrm{fg}}}\sum_{q\in\mathrm{fg}}
  \rho_q\langle\vtau_{b,q},\mW_q\vtau_{d,q}\rangle.
\]
We divide it by its mean diagonal,
\[
  \kappa(\vz)=\frac{\operatorname{tr}(\mK^{\mathrm{raw}}(\vz))}{n},
  \qquad
  \mK(\vz)=\frac{\mK^{\mathrm{raw}}(\vz)}{\kappa(\vz)}.
\]
This provides a common scale for the identity regularizer. 
At inference, we solve $(\mK(\vz)+\mI_n)\hat{\vs}=\vell(\vz)$ and subsequently set $\hat m_b=\mathbbm{1}[\hat s_b>0]$. 
This is an approximate correction: $\mK$ models the filtered-color branch rather than the complete nonlinear reader. 
Training supervises the raw scores $\vell$, without the joint solve.

\subsection{Embedding and publication}
\label{sec:supp:publication}
For embedding, the clean-host scores are evaluated on the canonical source image and treated as constants in the compensation branch. 
The remaining encoder and reader parameters are trained through the message-recovery loss.
After carrier mixing and upsampling, the residual is mapped back to the source-image crop, restricted to the source foreground, and made zero-mean over that foreground while remaining zero outside it. 
Let $\vr_x$ denote this source-frame residual. Publication uses
\[
  \vx'=Q_8\!\left(\operatorname{clip}_{[0,1]}
  (\vx+\alpha\vr_x)\right).
\]
The scalar strength $\alpha$ targets $35\,\mathrm{dB}$ foreground PSNR between $\vx$ and the clipped image \emph{before} $Q_8$. 
Thus, $35\,\mathrm{dB}$ denotes the pre-rounding target; the final published image includes 8-bit quantization.

\subsection{Training channels}
\label{sec:supp:channels}
\paragraph{\method-2D: No reconstructor in the loop.}
Each of the $V=4$ copies of $\vx'$ independently receives one of the following nine transformation types with equal probability: identity, JPEG compression (quality uniform in $\{50,\dots,95\}$), Gaussian blur ($\sigma\in[0.5,2]$ pixels), in-plane rotation ($[-10^\circ,10^\circ]$), a square crop keeping $[0.7,1]$ of the area, Gaussian noise ($\sigma\in[0.01,0.04]$), resizing (bicubic scale $[0.5,1.25]$), a horizontal flip, or a color change (brightness $\times[0.8,1.2]$, contrast $[0.8,1.2]$, saturation $[0.7,1.3]$, hue shift within $\pm0.03$ of the hue circle).

\paragraph{\method: TripoSR in the loop.}
Frozen TripoSR~\cite{tochilkin2024triposr} processes $\vx'$ using its $512\times512$ input preprocessing. With probability $0.5$, we jitter the input frame by shifts up to $\pm8$ pixels, scaling by up to $\pm4\%$, and blur with $\sigma\leq0.7$ pixels.
We extract a mesh using marching cubes on a $256^3$ grid and render it at $512\times512$ with nvdiffrast~\cite{laine2020nvdiffrast}, using unlit vertex colors, antialiasing, and a gray background. Each source image yields four views: one yaw sampled uniformly within each quadrant, with elevation uniform in $[-30^\circ,30^\circ]$.
Geometry is recomputed from the marked input in the forward pass and held constant in the backward pass; gradients propagate through vertex colors and rendering. TripoSR's parameters remain frozen. These training views differ from the primary evaluation range of yaw $\pm45^\circ$ and elevation $\pm15^\circ$.

\subsection{Optimization and schedule}
\label{sec:supp:optim}
We train the two variants independently from scratch using Adam with learning rate $10^{-4}$, gradient clipping at norm $1$ per parameter group, and minibatches of $8$ source images with $V=4$ channel outputs each. Messages are uniformly random, with a reserved pool for testing.
As in the main paper, binary cross-entropy supervises the raw scores on the channel outputs and marked source image, with source-term weight $\beta=0.5$. The trainable components are the carrier network ($3.96$\,M parameters), the evidence extractor, and the spatial-weight predictor with the mixing scalars and $\lambda$ ($3.2$\,k). The feature backbones remain frozen.
Each variant is trained for $3{,}000$ steps, taking $5.3$\,h for \method and $3.1$\,h for \method-2D on an NVIDIA RTX 4090. Checkpoint selection uses validation objects.
For the image-independent ablation, we replace the carrier network with $32$ learned RGB fields on the $64\times64$ grid, shared across all images. This variant has $393{,}216$ carrier parameters, compared with $3.96$\,M for the conditioned mapping; it therefore changes parameter count as well as content dependence.

\subsection{Data}
\label{sec:supp:data}

\begin{table}[t]
\centering
\caption{\textbf{Measured false-positive rates on unmarked renders} (\%). Mean averages the rates of $200$ random candidate messages; Max is the largest observed among those messages. The nominal random-message probabilities are $0.105\%$ at $k=25$ and $1.003\%$ at $k=23$. $\star$: the training reconstructor.}
\vspace{-3mm}
\label{tab:supp:fpr}
\small
\setlength{\tabcolsep}{3.2pt}
\begin{tabular}{lcccc}
\toprule
& \multicolumn{2}{c}{$k=25$ ($0.105\%$)} & \multicolumn{2}{c}{$k=23$ ($1.003\%$)} \\
\cmidrule(lr){2-3}\cmidrule(lr){4-5}
Reconstructor & Mean & Max & Mean & Max \\
\midrule
TripoSR$^\star$ & 0.10 & 0.46 & 0.95 & 3.67 \\
Real3D & 0.10 & 0.91 & 1.03 & 4.65 \\
OpenLRM & 0.11 & 1.33 & 0.98 & 2.83 \\
VFusion3D & 0.12 & 3.31 & 1.01 & 5.53 \\
SF3D & 0.11 & 1.06 & 1.06 & 5.78 \\
TGS & 0.11 & 1.17 & 1.03 & 4.96 \\
Splatter Image & 0.11 & 0.82 & 1.02 & 4.90 \\
TRELLIS & 0.11 & 1.25 & 0.98 & 5.56 \\
\bottomrule
\end{tabular}
\vspace{-3mm}
\end{table}

\begin{table}[t]
\centering
\caption{\textbf{Fixed-threshold detection and in-sample pooled calibration} (TPR, \%). Fixed uses $k=23$ or $k=25$. 
Calib.\ uses per-key thresholds fitted to unmarked renders from all eight reconstructors, targeting $1\%$ or $0.1\%$ pooled FPR on that same negative set. 
TPR is measured on separate marked test objects; the FPR targets are not independently validated test rates. 
$\star$: the training reconstructor.}
\vspace{-3mm}
\label{tab:supp:calib}
\small
\setlength{\tabcolsep}{4pt}
\resizebox{\columnwidth}{!}{%
\begin{tabular}{lcccc}
\toprule
& \multicolumn{2}{c}{$k=23$ / target $1\%$}
& \multicolumn{2}{c}{$k=25$ / target $0.1\%$} \\
\cmidrule(lr){2-3}\cmidrule(lr){4-5}
Reconstructor & Fixed & Calib. & Fixed & Calib. \\
\midrule
TripoSR$^\star$ & 90.0 & 88.4 & 80.6 & 79.4 \\
Real3D & 92.2 & 91.1 & 84.6 & 83.1 \\
OpenLRM & 91.9 & 90.9 & 85.6 & 83.5 \\
VFusion3D & 86.8 & 85.8 & 79.2 & 77.9 \\
SF3D & 94.0 & 93.6 & 89.5 & 88.9 \\
TGS & 81.2 & 79.6 & 67.6 & 63.8 \\
Splatter Image & 88.8 & 86.1 & 78.1 & 75.4 \\
TRELLIS & 42.6 & 39.9 & 25.5 & 23.5 \\
\midrule
Mean all 8 & 83.4 & 81.9 & 73.9 & 71.9 \\
\bottomrule
\end{tabular}%
}
\vspace{-3mm}
\end{table}

The main experiment uses $20{,}000$ Objaverse~\cite{deitke2023objaverse} objects for training, $500$ for validation, and $500$ for testing. 
We render one centered, scaled $512\times512$ RGBA source image per object using Blender Cycles, a transparent background, and uniform white environment lighting. 
Cross-dataset evaluation uses $250$ GSO~\cite{downs2022gso} and $250$ Toys4K~\cite{stojanov2021toys4k} objects, with one existing RGBA source image per object.

\section{Measured false-positive rates}
\label{sec:supp:fpr}

\begin{figure*}[t]
  \centering
  \includegraphics[width=\textwidth]{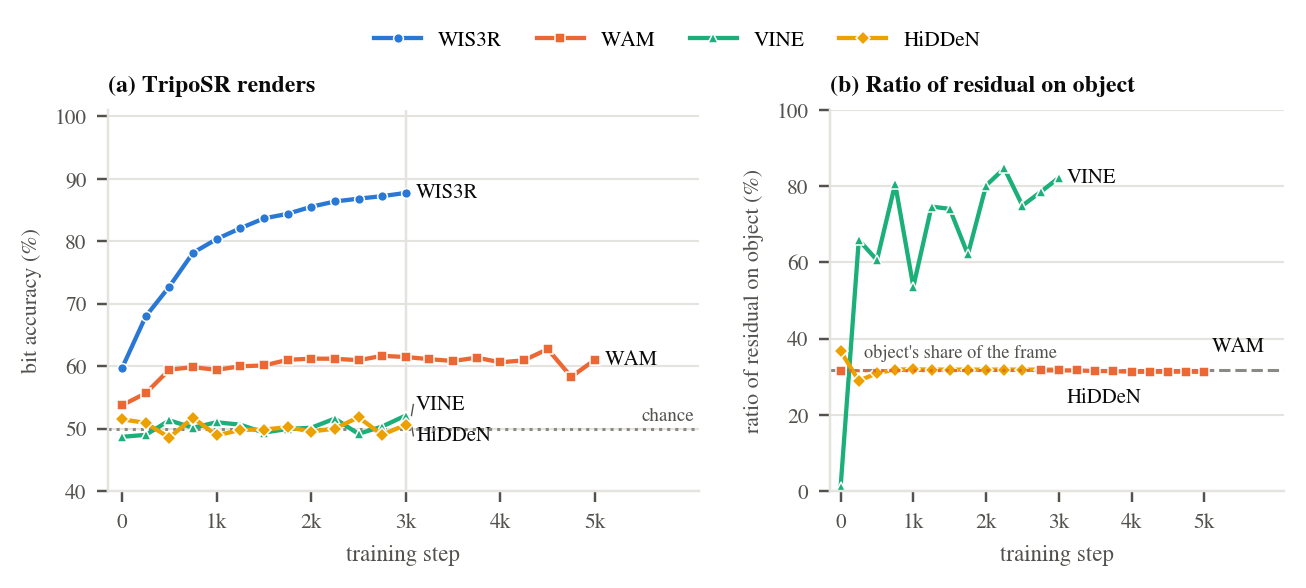}
  \vspace{-9mm}
 \caption{\textbf{Fine-tuning through TripoSR.} Left: reconstruction-channel bit accuracy as a function of fine-tuning step. 
 Right: the fraction of each baseline encoder's raw residual inside the foreground, measured \emph{before foreground masking}; the dashed line marks the object's share of the image area. 
 The residual is subsequently masked for publication. 
 WAM is trained for $5{,}000$ steps, and the other runs for $3{,}000$.}
  \label{fig:supp:ft_curves}
  \vspace{-3mm}
\end{figure*}

\begin{table*}[t]
\centering
\caption{\textbf{Fine-tuning configurations and recovery metrics.} All methods use the TripoSR channel, $32$ evaluated bits, and a $35\,\mathrm{dB}$ foreground target. 
Steps gives the training budget; bit accuracy and TPRs are percentages.}
\vspace{-3mm}
\label{tab:supp:finetune}
\small
\setlength{\tabcolsep}{4pt}
\begin{tabular}{llcccc}
\toprule
Method & Recipe and schedule & Steps & Bit acc. & $\mathrm{TPR}_{25}$ & $\mathrm{TPR}_{23}$ \\
\midrule
VINE~\cite{lu2025vine} & learning rate $10^{-4}$ constant, frozen U-Net & $3{,}000$ & 51.3 & 1.6 & 1.6 \\
WAM~\cite{sander2025wam} & learning rate $10^{-4}$ constant, no JND attenuation & $5{,}000$ & 62.7 & 5.9 & 21.5 \\
HiDDeN~\cite{zhu2018hidden} & learning rate $10^{-3}$ constant & $3{,}000$ & 51.1 & 0.0 & 1.6 \\
\midrule
\method & \cref{sec:supp:optim} & $3{,}000$ & 86.2 & 80.6 & 90.0 \\
\bottomrule
\end{tabular}
\vspace{-3mm}
\end{table*}

Let $C(\hat{\vm},\vm)$ count matching bits between the decoder output $\hat{\vm}$ and a candidate owner message $\vm$. 
Bit accuracy is $C/n$, and detection at threshold $k$ requires $C\geq k$. The decoder does not receive the candidate message.
For a uniformly random $n$-bit candidate independent of the decoded output, $C\sim\operatorname{Binomial}(n,1/2)$, regardless of the output distribution. 
The nominal false-match probability is
\begin{equation}
  p_k=2^{-n}\sum_{j=k}^{n}\binom{n}{j}.
  \label{eq:nominal_fpr}
\end{equation}
For $n=32$, $p_{25}=0.105\%$ and $p_{23}=1.003\%$. 
These probabilities average over random candidate messages; they do not guarantee the same FPR for every fixed owner message. 
We retain the main paper's fixed-threshold notation $\mathrm{TPR}_{25}$ and $\mathrm{TPR}_{23}$.

\paragraph{Measured per-key false positives.}
We reconstruct $1{,}000$ unmarked objects, disjoint from the watermark-training, validation, and marked-test objects. 
We evaluate eight renders per object and reconstructor against $200$ uniformly sampled candidate messages. 
\Cref{tab:supp:fpr} reports the mean per-key FPR and the largest observed FPR among those $200$ messages. 
The means are close to the nominal values, as expected when averaging over random candidates. 
The sampled maxima reach $3.31\%$ at $k=25$ and $5.78\%$ at $k=23$, demonstrating substantial variation across owner messages. 
These are maxima over the sampled messages, rather than guarantees over the entire key space.

\paragraph{Pooled per-key calibration.}
We additionally estimate one threshold per owner message using unmarked renders pooled across all eight reconstructors: $1{,}000$ objects $\times\,8$ reconstructors $\times\,8$ views, or $64{,}000$ renders. 
The threshold is shared across reconstructors at reading time, and the encoder and decoder remain fixed. 
The per-key columns of \cref{tab:supp:calib} report TPRs with calibration targets of $1\%$ and $0.1\%$.

The thresholds are fitted and their FPRs measured on the same $64{,}000$ unmarked renders; no separate negative test set is used for this calibration analysis. 
The marked test objects used to measure TPR are disjoint from these negative objects. 
These are therefore in-sample pooled calibration operating points, rather than independently validated FPRs. 
Moreover, the calibration pool includes outputs from the seven reconstructors excluded from watermark training. 
We report this analysis separately from the main paper's fixed-threshold zero-shot results; a pooled calibration target does not guarantee the same FPR for each reconstructor.

\section{Baselines}
\label{sec:supp:baselines}

\begin{table*}[t]
\centering
\caption{\textbf{Reconstruction pipelines and evaluation renderers.} Architecture and representation overview for the eight pipelines used in the main evaluation.}
\vspace{-3mm}
\label{tab:supp:i23d}
\small
\setlength{\tabcolsep}{4pt}
\begin{tabular}{lll}
\toprule
Reconstructor & Pipeline & Renderer \\
\midrule
TripoSR~\cite{tochilkin2024triposr} & DINO ViT-B/16 $\to$ transformer $\to$ triplane NeRF $\to$ mesh & nvdiffrast (mesh) \\
Real3D~\cite{jiang2024real3d} & TripoSR architecture, trained on real images & volume rendering \\
OpenLRM-1.1~\cite{he2023openlrm} & DINOv2 ViT-B/14 $\to$ transformer $\to$ triplane NeRF & volume rendering \\
VFusion3D~\cite{han2024vfusion3d} & DINOv2 ViT-B/14 $\to$ transformer $\to$ triplane NeRF & volume rendering \\
SF3D~\cite{boss2024sf3d} & DINOv2 ViT-L/14 $\to$ transformer $\to$ triplane $\to$ textured mesh & nvdiffrast (mesh) \\
TGS~\cite{zou2024tgs} & DINOv2 ViT-B/14 $\to$ point cloud + triplane $\to$ 3D Gaussians & gsplat \\
Splatter Image~\cite{szymanowicz2024splatter} & U-Net $\to$ one 3D Gaussian per pixel & gsplat \\
TRELLIS~\cite{xiang2025trellis} & DINOv2 ViT-L/14 $\to$ flow transformers $\to$ structured latent $\to$ mesh & nvdiffrast (mesh) \\
\bottomrule
\end{tabular}
\vspace{-3mm}
\end{table*}

\subsection{Released watermarkers}
\label{sec:supp:released}
Each released neural encoder receives the same canonical object frame as \method (\cref{sec:supp:frame}). 
We restrict its residual to the foreground and rescale it to the same $35\,\mathrm{dB}$ pre-rounding foreground PSNR target, after clipping. 
Each method retains its released encoder, decoder, and native payload length, as listed in the main paper. 
DWT-DCT uses its native strength and is not distortion matched.

\subsection{Fine-tuning through TripoSR}
\label{sec:supp:finetune}
We fine-tune VINE, WAM, and HiDDeN through the same TripoSR reconstruction and rendering channel used by \method, with the same foreground preprocessing and $35\,\mathrm{dB}$ target. 
We evaluate $32$ message positions and fix unused positions when a baseline has a larger native payload. 
Checkpoint selection uses validation objects, as described in the main paper.

\Cref{tab:supp:finetune} lists the reported recipes and recovery metrics. 
We also tested WAM with its JND attenuation enabled and disabled, obtaining $58.4\%$ and $62.7\%$ bit accuracy, respectively, and explored multiple learning rates for VINE and HiDDeN. 
The runs in \cref{fig:supp:ft_curves} cover $3{,}000$ steps for \method, VINE, and HiDDeN; WAM is extended to $5{,}000$ steps.

\section{Additional results}
\label{sec:supp:results}

We expand the main evaluation with wider-view detection rates (\cref{sec:supp:views}), source-image fidelity comparisons (\cref{sec:supp:fidelity}), reconstruction-pipeline details (\cref{sec:supp:i23d}), and estimated foreground masks (\cref{sec:supp:rmbg}).

\subsection{Viewpoint regions}
\label{sec:supp:views}
\Cref{tab:supp:views} groups renders by absolute yaw relative to the source view. 
The unseen-model mean bit accuracy is $83.7\%$ in the front region, $68.8\%$ on the near side, $63.5\%$ on the far side, and $67.5\%$ at the rear. 
Recovery is therefore not monotonic in yaw; these measurements do not isolate the cause of the rear-view increase.
Bit accuracy remains above chance in all four regions, but message detection declines substantially: unseen-model $\mathrm{TPR}_{25}$ falls from $72.9\%$ in the front region to $11.7\%$ on the far side. 
The results support partial message retention at large viewpoint changes, with the most reliable detection near the conditioning view.

\begin{table}[H]
\centering
\caption{\textbf{Recovery across viewpoint regions} (\%). 
Yaw is measured relative to the source view; elevation lies in $[-15^\circ,15^\circ]$. 
Unseen seven averages the seven main-paper pipelines excluding TripoSR. 
TPRs use the fixed match-count thresholds.}
\vspace{-3mm}
\label{tab:supp:views}
\small
\setlength{\tabcolsep}{3pt}
\resizebox{\columnwidth}{!}{%
\begin{tabular}{llcccccc}
\toprule
& & \multicolumn{3}{c}{TripoSR} & \multicolumn{3}{c}{Unseen seven} \\
\cmidrule(lr){3-5}\cmidrule(lr){6-8}
Region & Absolute yaw & Bit acc. & $\mathrm{TPR}_{25}$ & $\mathrm{TPR}_{23}$ & Bit acc. & $\mathrm{TPR}_{25}$ & $\mathrm{TPR}_{23}$ \\
\midrule
Front & $0\dg$--$45\dg$ & 86.2 & 80.6 & 90.0 & 83.7 & 72.9 & 82.5 \\
Near side & $45\dg$--$90\dg$ & 71.4 & 34.0 & 57.5 & 68.8 & 27.1 & 44.9 \\
Far side & $90\dg$--$135\dg$ & 64.7 & 11.0 & 31.7 & 63.5 & 11.7 & 26.7 \\
Rear & $135\dg$--$180\dg$ & 69.4 & 27.9 & 45.0 & 67.5 & 23.7 & 40.1 \\
\bottomrule
\end{tabular}%
}
\vspace{-3mm}
\end{table}

\subsection{Published-image fidelity}
\label{sec:supp:fidelity}
\Cref{tab:supp:fidelity} compares marked and clean source images on the foreground at the common $35\,\mathrm{dB}$ pre-rounding target. 
Both \method variants achieve SSIM $0.990$, compared with $0.965$ for fine-tuned WAM. 
\method's LPIPS is $0.045$, compared with $0.084$ for fine-tuned WAM and $0.064$ for released WAM.
These metrics evaluate the published source images; they do not measure changes to the reconstructed assets.

\begin{table}[H]
\centering
\caption{\textbf{Published source-image fidelity.} Foreground SSIM and LPIPS at the common $35\,\mathrm{dB}$ pre-rounding PSNR target. 
Lower LPIPS and higher SSIM indicate closer agreement with the clean source image.}
\vspace{-3mm}
\label{tab:supp:fidelity}
\small
\setlength{\tabcolsep}{3pt}
\begin{tabular}{lcc}
\toprule
Method & SSIM $\uparrow$ & LPIPS $\downarrow$ \\
\midrule
\method & 0.990 & 0.045 \\
\method-2D & 0.990 & 0.042 \\
WAM, fine-tuned (\cref{tab:supp:finetune}) & 0.965 & 0.084 \\
WAM, released & 0.967 & 0.064 \\
Random noise & 0.921 & 0.055 \\
\bottomrule
\end{tabular}
\vspace{-3mm}
\end{table}

\subsection{Reconstructors}
\label{sec:supp:i23d}
\Cref{tab:supp:i23d} summarizes the reconstruction pipelines and evaluation renderers. 
We use released reconstruction weights, with the renderer listed for each pipeline. 
Among the eight main-paper pipelines, TripoSR and TRELLIS are rendered at $512\times512$; the other six are rendered at $256\times256$ and resized for canonical reading.
Real3D uses the TripoSR architecture with real-image training. 
Its exclusion from watermark training establishes transfer to that released model, while the other pipelines provide additional architectural diversity.

\subsection{Estimated foreground masks}
\label{sec:supp:rmbg}
Unless stated otherwise, rendered-image experiments use the renderer's coverage as the foreground mask. 
To evaluate reading without this mask, we estimate foreground directly from each received render using RMBG-2.0.\footnote{\url{https://huggingface.co/briaai/RMBG-2.0}}
Recovery is comparable with estimated masks (\cref{tab:supp:rmbg}): bit accuracy is $86.2\%$ on TripoSR and $83.9\%$ on the unseen seven, compared with $86.2\%$ and $83.7\%$ using coverage masks.
Mean mask IoU ranges from $0.93$ to $0.99$ across reconstructors. 
Estimating masks at both publication and reading yields $86.6\%$ bit accuracy and $81.6\%$ $\mathrm{TPR}_{25}$ on TripoSR, and $84.0\%$ bit accuracy and $73.4\%$ $\mathrm{TPR}_{25}$ over the seven unseen reconstructors.

\begin{table}[H]
\centering
\caption{\textbf{Estimated foreground masks.} Bit accuracy and fixed-threshold $\mathrm{TPR}_{25}$ on object-centric renders. 
The encoder and reader weights remain fixed across the mask conditions.}
\vspace{-3mm}
\label{tab:supp:rmbg}
\setlength{\tabcolsep}{3pt}
\resizebox{\columnwidth}{!}{%
\begin{tabular}{lcccc}
\toprule
& \multicolumn{2}{c}{TripoSR} & \multicolumn{2}{c}{Unseen seven} \\
\cmidrule(lr){2-3}\cmidrule(lr){4-5}
Condition & Bit acc. & $\mathrm{TPR}_{25}$ & Bit acc. & $\mathrm{TPR}_{25}$ \\
\midrule
Renderer coverage mask & 86.2\% & 80.6\% & 83.7\% & 72.9\% \\
RMBG-2.0 mask at read time & 86.2\% & 81.1\% & 83.9\% & 73.3\% \\
RMBG-2.0 masks at publication and read time & 86.6\% & 81.6\% & 84.0\% & 73.4\% \\
\bottomrule
\end{tabular}%
}
\vspace{-3mm}
\end{table}

\begin{table*}[h!]
\caption{\textbf{\method encoder operations for $n=32$ bits.} Strength is selected after clipping and before 8-bit rounding.}
    \label{tab:wis3r_encoder}
    \begin{center}
        \small
        \setlength{\tabcolsep}{3pt}
        \begin{tabular}{*{3}{l}}
        \toprule
        & \textbf{WIS3R encoder architecture} & $\leftarrow\mathit{Encode}(\vx,\mM,\vm)$ \\
        \midrule
        Index & Layer & Out \\
        \midrule
        1 & \textit{Input}: source image $\vx$, foreground mask $\mM$, message $\vm$
          & $H\!\times\!W\!\times\!3$; $H\!\times\!W$; $32$ bits \\
        2 & $\mathit{Canonicalize}(\vx,\mM)$
          & $512\!\times\!512\!\times\!3$: $\vz$; $512\!\times\!512$: $\mM_z$ \\
        3 & Cell mask from $\mM_z$
          & $64\!\times\!64\!\times\!1$ \\
        4 & $\mathit{SharedFeatures}(\vz)$  (Table~\ref{tab:wis3r_shared})
          & $64\!\times\!64\!\times\!3200$: $F(\vz)$ \\
        5 & $\mathit{Carriers}(\mathrm{Out}(4),\mM_g)$  (Table~\ref{tab:wis3r_shared})
          & $32\!\times\!64\!\times\!64\!\times\!3$: $\vtau(\vz)$ \\
        6 & $\hat{\vh}=\mathit{stopgrad}\!\left(\mathit{RawRead}(\vz,\mM_z)_{\vell}\right)$
          & $32$: clean-host scores $\hat{\vh}$ \\
        7 & $\vs=2\vm-1$
          & $32$: message signs $\vs$ \\
        8 & $\mathrm{Out}(7)-\lambda\,\mathrm{Out}(6)$
          & $32$: coefficients $\vc$ \\
        9 & $\displaystyle\sum_{b=1}^{32}\mathrm{Out}(8)_b\,\mathrm{Out}(5)_b$
          & $64\!\times\!64\!\times\!3$: carrier mixture \\
        10 & $\mathcal{U}\!\left(\mathrm{Out}(9);\,64\!\times\!64\rightarrow512\!\times\!512\right)$
           & $512\!\times\!512\!\times\!3$ \\
        11 & $\mM_z\odot\mathrm{Out}(10)$
           & $512\!\times\!512\!\times\!3$: residual $\vr$ \\
        12 & $\mM\odot\mathcal{B}_x\!\left(\mathrm{Out}(11)\right)$
           & $H\!\times\!W\!\times\!3$: source-frame residual \\
        13 & $\mM\odot\left[\mathrm{Out}(12)-\vmu_{\mM}\!\left(\mathrm{Out}(12)\right)\right]$
           & $H\!\times\!W\!\times\!3$: centered residual $\vr_x$ \\
        14 & Choose $\alpha$ such that $\mathit{PSNR}_{\mM}\!\left(\vx,\mathit{clip}_{[0,1]}(\vx+\alpha \vr_x)\right)=35\,\mathrm{dB}$
           & $1$: embedding strength $\alpha$ \\
        15 & $\mathit{clip}_{[0,1]}\!\left(\vx+\mathrm{Out}(14)\,\mathrm{Out}(13)\right)$
           & $H\!\times\!W\!\times\!3$: clipped image \\
        16 & $Q_8\!\left(\mathrm{Out}(15)\right)$
           & $H\!\times\!W\!\times\!3$: published image $\vx'$ \\
        \bottomrule
        \end{tabular}
    \end{center}
\end{table*}

\begin{table*}[h!]
    \caption{\textbf{Raw reader and joint decoder.} Spatial weights and carrier overlap are normalized before the inference-only joint solve.}
    \label{tab:wis3r_decoder}
    \begin{center}
        \small
        \setlength{\tabcolsep}{3pt}
        \begin{tabular}{*{3}{l}}
        \toprule
        & \textbf{WIS3R decoder architecture} & $\leftarrow\mathit{Decode}(\vy,\mM_y)$ \\
        \midrule
        Index & Layer & Out \\
        \midrule
        1 & \textit{Input}: received image $\vy$, foreground mask $\mM_y$
          & $H_y\!\times\!W_y\!\times\!3$; $H_y\!\times\!W_y$ \\
        2 & $\mathit{Canonicalize}(\vy,\mM_y)$
          & $512\!\times\!512\!\times\!3$: $\vz$; $512\!\times\!512$: $\mM_z$ \\
        3 & Cell mask from $\mM_z$
          & $64\!\times\!64\!\times\!1$ \\
        4 & $\mathit{SharedFeatures}(\vz)$  (Table~\ref{tab:wis3r_shared})
          & $64\!\times\!64\!\times\!3200$: $F(\vz)$ \\
        5 & $\mathit{Carriers}(\mathrm{Out}(4),\mM_g)$  (Table~\ref{tab:wis3r_shared})
          & $32\!\times\!64\!\times\!64\!\times\!3$: $\vtau(\vz)$ \\
        6 & $\mathcal{D}(\vz;\mM_z)$: average foreground pixels within each cell
          & $64\!\times\!64\!\times\!3$: cell colors $\vd$ \\
        7 & $\mathcal{H}(\mathrm{Out}(6);\mM_g)$ \quad ($\sigma=6$ cells)
          & $64\!\times\!64\!\times\!3$: color detail $\vu$ \\
        8 & $\mSigma_q=\mathcal{G}_{4}^{\mathrm{fg}}\!\left[\vu \vu^{\mathsf T}\right]_q$
          & $64\!\times\!64\!\times\!3\!\times\!3$: $\mSigma$ \\
        9 & $\displaystyle\bar{\sigma}^{2}=\frac{1}{N_{\mathrm{fg}}}\sum_{q\in\mathrm{fg}}\frac{\operatorname{tr}(\mSigma_q)}{3}$
          & $1$: mean color variance \\
        10 & $\mW_q=\bar{\sigma}^{2}\!\left(\mSigma_q+0.03\,\bar{\sigma}^{2}\mI_3\right)^{-1}$
           & $64\!\times\!64\!\times\!3\!\times\!3$: $\mW$ \\
        11 & $\widetilde{\vu}_q=\mathrm{Out}(10)_q\,\mathrm{Out}(7)_q$
           & $64\!\times\!64\!\times\!3$: weighted detail $\widetilde{\vu}$ \\
        12 & $\mathit{Extractor}\!\left(\mathrm{Out}(6),\mathrm{Out}(11),\mathrm{Out}(4),\mM_g\right)$ (Table~\ref{tab:wis3r_blocks})
           & $64\!\times\!64\!\times\!3$: evidence $\widehat{\vdelta}$ \\
        13 & $a_1\,\mathrm{Out}(11)+a_2\,\mathrm{Out}(12)$
           & $64\!\times\!64\!\times\!3$: combined evidence $\vv$ \\
        14 & $\mathit{Softplus}\!\left(\mathit{Conv}_{1\times1}(\mathrm{Out}(4);3200\rightarrow1)\right)$
           & $64\!\times\!64\!\times\!1$: positive weights $\rho^+$ \\
        15 & $\displaystyle\rho_q=\rho_q^+\bigg/\left(\frac{1}{N_{\mathrm{fg}}}\sum_{p\in\mathrm{fg}}\rho_p^+\right)$
           & $64\!\times\!64\!\times\!1$: normalized weights $\rho$ \\
        16 & $e_{b,q}=\rho_q\left\langle\vtau_{b,q},\vv_q\right\rangle$
           & $32\!\times\!64\!\times\!64$: cell contributions \\
        17 & $\displaystyle\ell_b=\frac{1}{N_{\mathrm{fg}}}\sum_{q\in\mathrm{fg}}\mathrm{Out}(16)_{b,q}$
           & $32$: raw scores $\vell$ \\
        18 & $\displaystyle K^{\mathrm{raw}}_{bd}=\frac{1}{N_{\mathrm{fg}}}\sum_{q\in\mathrm{fg}}\rho_q\left\langle\vtau_{b,q},\mW_q\vtau_{d,q}\right\rangle$
           & $32\!\times\!32$: raw overlap $\mK^{\mathrm{raw}}$ \\
        19 & $\kappa=\operatorname{tr}\!\left(\mathrm{Out}(18)\right)/32$
           & $1$: mean diagonal $\kappa$ \\
        20 & $\mK=\mathrm{Out}(18)/\mathrm{Out}(19)$
           & $32\!\times\!32$: normalized overlap $\mK$ \\
        \midrule
        \multicolumn{3}{c}{\textit{Joint decoding (inference only)}} \\
        \midrule
        21 & $\widehat{\vs}=\mathit{Solve}\!\left(\mathrm{Out}(20)+\mI_{32},\mathrm{Out}(17)\right)$
           & $32$: estimated signs $\widehat{\vs}$ \\
        22 & $\widehat{m}_b=\mathbf{1}\!\left[\mathrm{Out}(21)_b>0\right]$
           & $32$ bits: recovered message $\widehat{\vm}$ \\
        \bottomrule
        \end{tabular}
    \end{center}
\end{table*}
\clearpage

\begin{table*}[h!]
\caption{\textbf{Shared features and carrier construction.} The same weights and fixed channel statistics are used for embedding and reading.}
    \label{tab:wis3r_shared}
    \begin{center}
        \small
        \setlength{\tabcolsep}{3pt}
        \begin{tabular}{*{3}{l}}
        \toprule
        & \textbf{Shared feature backbone} & $\leftarrow\mathit{SharedFeatures}(\vz)$ \\
        \midrule
        Index & Layer & Out \\
        \midrule
        1 & \textit{Input}: canonical RGB image $\vz$
          & $512\!\times\!512\!\times\!3$ \\
        2 & Frozen DINO ViT-B/16: spatial tokens from layers $4,8,12$
          & $3\times(32\!\times\!32\!\times\!768)$ \\
        3 & Bilinear resize of each feature map in $\mathrm{Out}(2)$ to $64\!\times\!64$
          & $3\times(64\!\times\!64\!\times\!768)$ \\
        4 & $\mathit{Concat}_{\mathrm{ch}}\!\left(\mathrm{Out}(3)\right)$
          & $64\!\times\!64\!\times\!2304$ \\
        5 & Bicubic resize of $\vz$ to $256\!\times\!256$
          & $256\!\times\!256\!\times\!3$ \\
        6 & Frozen VGG16: $\texttt{relu2\_2}$ from $\mathrm{Out}(5)$
          & $128\!\times\!128\!\times\!128$ \\
        7 & Average-pool $\mathrm{Out}(6)$ to $64\!\times\!64$
          & $64\!\times\!64\!\times\!128$ \\
        8 & Frozen VGG16: $\texttt{relu3\_3}$ from $\mathrm{Out}(5)$
          & $64\!\times\!64\!\times\!256$ \\
        9 & Frozen VGG16: $\texttt{relu4\_3}$ from $\mathrm{Out}(5)$
          & $32\!\times\!32\!\times\!512$ \\
        10 & Bilinear resize of $\mathrm{Out}(9)$ to $64\!\times\!64$
           & $64\!\times\!64\!\times\!512$ \\
        11 & $\mathit{Concat}_{\mathrm{ch}}\!\left(\mathrm{Out}(4),\mathrm{Out}(7),\mathrm{Out}(8),\mathrm{Out}(10)\right)$
           & $64\!\times\!64\!\times\!3200$: $F_0(\vz)$ \\
        12 & Repeat rows~2--11 on $\mathit{flip}(\vz)$; flip the resulting grid back
           & $64\!\times\!64\!\times\!3200$: aligned mirror features \\
        13 & $\tfrac{1}{2}\left(\mathrm{Out}(11)+\mathrm{Out}(12)\right)$
           & $64\!\times\!64\!\times\!3200$ \\
        14 & Channelwise standardization: $\left(\mathrm{Out}(13)-\vmu_F\right)/\bm{\sigma}_F$
           & $64\!\times\!64\!\times\!3200$: $F(\vz)$ \\
        \midrule\midrule
        & \textbf{Carrier network and normalization} & $\leftarrow\mathit{Carriers}(\mF,\mM_g)$ \\
        \midrule
        Index & Layer & Out \\
        \midrule
        1 & \textit{Input}: shared features $\mF$, foreground-cell mask $\mM_g$
          & $64\!\times\!64\!\times\!3200$; $64\!\times\!64\!\times\!1$ \\
        2 & $\mathit{Conv}_{1\times1}(\mF;3200\rightarrow1024)$
          & $64\!\times\!64\!\times\!1024$ \\
        3 & $\mathit{SiLU}\!\left(\mathit{LayerNorm}_{\mathrm{ch}}(\mathrm{Out}(2))\right)$
          & $64\!\times\!64\!\times\!1024$ \\
        4 & $\mathit{Conv}_{1\times1}(\mathrm{Out}(3);1024\rightarrow512)$
          & $64\!\times\!64\!\times\!512$ \\
        5 & $\mathit{SiLU}\!\left(\mathit{LayerNorm}_{\mathrm{ch}}(\mathrm{Out}(4))\right)$
          & $64\!\times\!64\!\times\!512$ \\
        6 & $\mathit{Conv}_{1\times1}(\mathrm{Out}(5);512\rightarrow256)$
          & $64\!\times\!64\!\times\!256$ \\
        7 & $\mathit{SiLU}\!\left(\mathit{LayerNorm}_{\mathrm{ch}}(\mathrm{Out}(6))\right)$
          & $64\!\times\!64\!\times\!256$ \\
        8 & $\mathit{Conv}_{1\times1}(\mathrm{Out}(7);256\rightarrow96)$
          & $64\!\times\!64\!\times\!96$: $f_\theta(\mF)$ \\
        9 & Reshape and permute $\mathrm{Out}(8)$: $96=32\times3$
          & $32\!\times\!64\!\times\!64\!\times\!3$: $\vt$ \\
        10 & Apply $\mathcal{H}(\cdot;\mM_g)$ separately to each RGB field in $\mathrm{Out}(9)$
           & $32\!\times\!64\!\times\!64\!\times\!3$: $\mT$ \\
        11 & $\displaystyle r_b=\sqrt{\frac{1}{3N_{\mathrm{fg}}}\sum_{q\in\mathrm{fg}}\sum_{j=1}^{3}T_{b,q,j}^{\,2}}$
           & $32\!\times\!1\!\times\!1\!\times\!1$: per-carrier RMS \\
        12 & $\vtau=\mathrm{Out}(10)/\mathrm{Out}(11)$ \quad (broadcast division)
           & $32\!\times\!64\!\times\!64\!\times\!3$: carriers $\vtau$ \\
        \bottomrule
        \end{tabular}
    \end{center}
\end{table*}

\begin{table*}[h!]
    \caption{\textbf{Evidence extractor and basic operations.} Channel concatenation combines features at the same spatial resolution. 
    $\operatorname{SymConv}$ denotes a convolution with left--right symmetric kernels.}
    \label{tab:wis3r_blocks}
    \begin{center}
        \small
        \setlength{\tabcolsep}{3pt}
        \begin{tabular}{*{3}{l}}
        \toprule
        & \textbf{Watermark-evidence extractor} & $\leftarrow\mathit{Extractor}(\vd,\widetilde{\vu},\mF,\mM_g)$ \\
        \midrule
        Index & Layer & Out \\
        \midrule
        1 & \textit{Input}: cell colors $\vd$, weighted detail $\widetilde{\vu}$,
            shared features $\mF$, mask $\mM_g$
          & $2\times(64\!\times\!64\!\times\!3)$; $64\!\times\!64\!\times\!3200$; $64\!\times\!64\!\times\!1$ \\
        2 & $\mathit{Project48}(\mF)$\textsuperscript{$\dagger$}
          & $64\!\times\!64\!\times\!48$ \\
        3 & $\mathit{Concat}_{\mathrm{ch}}\!\left(\vd,\widetilde{\vu},\mathrm{Out}(2),\mM_g\right)$
          & $64\!\times\!64\!\times\!55$ \\
        4 & $\mathit{ConvBlock}(\mathrm{Out}(3);55\rightarrow48)$
          & $64\!\times\!64\!\times\!48$: high-resolution skip \\
        5 & $\mathit{AvgPool}_{2\times2,s=2}(\mathrm{Out}(4))$\textsuperscript{$\dagger$}
          & $32\!\times\!32\!\times\!48$ \\
        6 & $\mathit{ConvBlock}(\mathrm{Out}(5);48\rightarrow96)$
          & $32\!\times\!32\!\times\!96$: middle-resolution skip \\
        7 & $\mathit{AvgPool}_{2\times2,s=2}(\mathrm{Out}(6))$\textsuperscript{$\dagger$}
          & $16\!\times\!16\!\times\!96$ \\
        8 & $\mathit{ConvBlock}(\mathrm{Out}(7);96\rightarrow192)$
          & $16\!\times\!16\!\times\!192$: bottleneck \\
        9 & Bilinear resize of $\mathrm{Out}(8)$ to $32\!\times\!32$
          & $32\!\times\!32\!\times\!192$ \\
        10 & $\mathit{Concat}_{\mathrm{ch}}\!\left(\mathrm{Out}(9),\mathrm{Out}(6)\right)$
           & $32\!\times\!32\!\times\!288$ \\
        11 & $\mathit{ConvBlock}(\mathrm{Out}(10);288\rightarrow96)$
           & $32\!\times\!32\!\times\!96$ \\
        12 & Bilinear resize of $\mathrm{Out}(11)$ to $64\!\times\!64$
           & $64\!\times\!64\!\times\!96$ \\
        13 & $\mathit{Concat}_{\mathrm{ch}}\!\left(\mathrm{Out}(12),\mathrm{Out}(4)\right)$
           & $64\!\times\!64\!\times\!144$ \\
        14 & $\mathit{ConvBlock}(\mathrm{Out}(13);144\rightarrow48)$
           & $64\!\times\!64\!\times\!48$ \\
        15 & $\mathit{Conv}_{1\times1}(\mathrm{Out}(14);48\rightarrow3)$
           & $64\!\times\!64\!\times\!3$: evidence $\widehat{\vdelta}$ \\
        \midrule\midrule
        & \textbf{Two-convolution U-Net block} & $\leftarrow\mathit{ConvBlock}(\mX;D_{\rm in}\rightarrow D_{\rm out})$ \\
        \midrule
        Index & Layer & Out \\
        \midrule
        1 & \textit{Input}: feature map $\mX$
          & $h\!\times\!w\!\times\!{D_{\rm in}}$ \\
        2 & $\mathit{SymConv}_{3\times3}(\mX;D_{\rm in}\rightarrow D_{\rm out})$
          & $h\!\times\!w\!\times\!{D_{\rm out}}$ \\
        3 & $\mathit{SiLU}\!\left(\mathit{GroupNorm}_{8}(\mathrm{Out}(2))\right)$
          & $h\!\times\!w\!\times\!{D_{\rm out}}$ \\
        4 & $\mathit{SymConv}_{3\times3}(\mathrm{Out}(3);D_{\rm out}\rightarrow D_{\rm out})$
          & $h\!\times\!w\!\times\!{D_{\rm out}}$ \\
        5 & $\mathit{SiLU}\!\left(\mathit{GroupNorm}_{8}(\mathrm{Out}(4))\right)$
          & $h\!\times\!w\!\times\!{D_{\rm out}}$ \\
        \midrule\midrule
        & \textbf{Foreground-restricted high-pass filter} & $\leftarrow\mathcal{H}(\mX;\mM_g)$ \\
        \midrule
        Index & Layer & Out \\
        \midrule
        1 & \textit{Input}: field $\mX$, foreground-cell mask $\mM_g$
          & $64\!\times\!64\!\times\!C$; $64\!\times\!64\!\times\!1$ \\
        2 & $\displaystyle \mL_q=\frac{\sum_p G_6(q-p)M_g(p)\mX_p}{\sum_p G_6(q-p)M_g(p)}$ \quad for $q\in\mathrm{fg}$
          & $64\!\times\!64\!\times\!C$: local mean $\mL$ \\
        3 & $\mathcal{H}[\mX]_q=\mX_q-\mL_q$ for $q\in\mathrm{fg}$; $0$ otherwise
          & $64\!\times\!64\!\times\!C$: high-pass field \\
        \bottomrule
        \end{tabular}
    \end{center}
\end{table*}

\typeout{get arXiv to do 4 passes: Label(s) may have changed. Rerun}

\end{document}